\documentclass[runningheads]{llncs}
\usepackage{makeidx}
\usepackage{graphicx}
\usepackage{array}
\usepackage{amsmath,amssymb} 
\usepackage{color}
\usepackage{caption}
\usepackage{subcaption}
\usepackage{ulem}
\usepackage{textcomp}
\usepackage{booktabs}
\usepackage{hyperref}
\usepackage{xcolor}
\hypersetup{
    colorlinks,
    linkcolor={red!50!black},
    citecolor={blue!50!black},
    urlcolor={blue!80!black}
}
\usepackage{fancyhdr}
\fancypagestyle{gcpr}{
	\fancyhf{}

	\fancyfoot[L]{
	\footnotesize{\vspace{-0.4cm}Preliminary version. Accepted for 42nd German Conference on Pattern Recognition (GCPR), 2020, T\"ubingen, Germany}
	}{}
}

\makeatletter
\renewcommand*{\eqref}[1]{%
  \hyperref[{#1}]{\textup{\tagform@{\ref*{#1}}}}%
}
\makeatother

\newcommand\blfootnote[1]{%
  \begingroup
  \renewcommand\thefootnote{}\footnote{#1}%
  \addtocounter{footnote}{-1}%
  \endgroup
}

\begin{document}
	\pagestyle{headings}
	\mainmatter

	\title{Learning to Identify Physical Parameters from Video Using Differentiable Physics}
	\titlerunning{Learning to Identify Physical Parameters Using Differentiable Physics}
	\authorrunning{Rama Kandukuri, J. Achterhold, Michael Moeller, Joerg Stueckler}
	\author{Rama Kandukuri$^{1,2}$, Jan Achterhold$^1$, Michael Moeller$^2$, Joerg Stueckler$^1$}
	\institute{\vspace{-0.2cm}$^1$Max Planck Institute for Intelligent Systems, T\"ubingen, Germany \\$^2$University of Siegen, Germany}
	\maketitle
	
	\thispagestyle{gcpr}

    \begin{abstract}
\vspace{-0.5cm}
Video representation learning has recently attracted attention in computer vision due to its applications for activity and scene forecasting or vision-based planning and control. Video prediction models often learn a latent representation of video which is encoded from input frames and decoded back into images. Even when conditioned on actions, purely deep learning based architectures typically lack a physically interpretable latent space. In this study, we use a differentiable physics engine within an action-conditional video representation network to learn a physical latent representation. We propose supervised and self-supervised learning methods to train our network and identify physical properties. The latter uses spatial transformers to decode physical states back into images. The simulation scenarios in our experiments comprise pushing, sliding and colliding objects, for which we also analyze the observability of the physical properties. In experiments we demonstrate that our network can learn to encode images and identify physical properties like mass and friction from videos and action sequences in the simulated scenarios. We evaluate the accuracy of our supervised and self-supervised methods and compare it with a system identification baseline which directly learns from state trajectories. We also demonstrate the ability of our method to predict future video frames from input images and actions.

	\end{abstract}

	\section{Introduction}
	\label{sec:introduction}
	\blfootnote{Corresponding author: Rama Kandukuri (\texttt{rama.kandukuri@tue.mpg.de})}
	Video representation learning is a challenging task in computer vision which has applications in scene 	understanding and prediction~\cite{motthagi2016_whatif,mottaghi2016_newtonian} or vision-based control and planning~\cite{finn2016_unsupvideopredact,finn2017_visualforesight,hafner2019_planet}. Such approaches can be distinguished into supervised or self-supervised methods, the latter typically based on recurrent autoencoder models which are trained for video prediction.
    
	Typical architectures of video prediction models first encode the image in a low dimensional latent scene representation.
	This latent state is predicted forward eventually based on actions and finally decoded into future frames. 
	Neural network based video prediction models like \cite{srivastava2015_unsupvideoreplstm,finn2016_unsupvideopredact,babaeizadeh2018_stochvarvideopred} perform these steps implicitly and typically learn a latent representation which cannot be directly interpreted for physical quantities such as mass, friction, position and velocity. 
	This can limit explainability and generalization for new tasks and scenarios. 
	Analytical models like \cite{analyticalLearned,degrave2019_diffphys,belbuteperes2018_diffphys} in contrast structure the latent space as an interpretable physical parameterization and use analytical physical models to forward the latent state. 

	In this paper we study supervised and self-supervised learning approaches for identifying physical parameters of objects from video.
	Our approach encodes images into physical states and uses a differentiable physics engine~\cite{belbuteperes2018_diffphys} to forward the physical scene state based on latent physical scene parameters. 
	For self-supervised learning, we apply spatial transformers~\cite{stn} to decode the predicted physical scene states into images based on known object models.
	We evaluate our approach in various simulation scenarios such as pushing, sliding and collision of objects and analyze the observability of physical parameters in these scenarios. 
	In our experiments, we demonstrate that physical scene encodings can be learned from video and interactions through supervised and self-supervised training.
	Our method allows for identifying the observable physical parameters of the objects from the videos.
		In summary, we make the following contributions in this work
	\begin{itemize}
	    \item We propose supervised and self-supervised learning approaches to learn to encode scenes into physical scene representations of objects. Our novel architecture integrates a differentiable physics engine as a forward model. 
	    It uses spatial transformers to decode the states back into images for self-supervised learning.
	    \item We analyse the observability of physical parameters in pushing, sliding and collision scenarios. 
	    Our approach simultaneously identifies the observable physical parameters during training while learning the network parameters of the encoder.
	    \item We evaluate our approach on simulated scenes and analyse its accuracy in recovering object pose and physical parameters.
	\end{itemize}

	\subsection{Related Work}
	\label{sec:related work}
	
    \subsubsection{Neural video prediction}
    Neural video prediction models learn an embedding of video frames into a latent representation using successive neural network operations such as convolutions, non-linearities and recurrent units.
    Srivastava et al.~\cite{srivastava2015_unsupvideoreplstm} embed images into a latent representation recurrently using long short term memory (LSTM~\cite{hochreiter1997_lstm}) cells. 
    The latent representation is decoded back using a convolutional decoder.
    Video prediction is achieved by propagating the latent representation of the LSTM forward using predicted frames as inputs.
    Finn et al.~\cite{finn2016_unsupvideopredact} also encode images into a latent representation using successive LSTM convolutions~\cite{shi2015_convlstm}.
    The decoder predicts motion kernels ($5\times 5$ pixels) and composition masks for the motion layers which are used to propagate the input images.
    
        A typical problem of such architectures is that they cannot capture multi-modal distributions on future frames well, for example, in the case of uncertain interactions of objects, which leads to blurry predictions.
    Babaeizadeh et al.~\cite{babaeizadeh2018_stochvarvideopred} introduce a stochastic latent variable which is inferred from the full sequence at training time and sampled from a fixed prior at test time.
    Visual interaction networks explicitly model object interactions using graph neural networks in a recurrent video prediction architecture~\cite{watters2017_vin}.
    However, these approaches do not learn a physically interpretable latent representation and cannot be used to infer physical parameters.
    To address this shortcomings, Ye et al.~\cite{ye2018_interppred} train a variational autoencoder based architecture in a conditional way by presenting training data with variation in each single specific physical property while holding all but a few latent variables fixed. 
    This way, the autoencoder is encouraged to represent this property in the corresponding part of the latent vector.
    The approach is demonstrated on videos of synthetic 3D scenes with colliding shape primitives.
    Zhu et al.~\cite{zhu2019_physvidpred} combine disentangled representation learning based on total correlation~\cite{chen2018_tcbetavae} with partial supervision of physical properties. 
    These purely deep learning based techniques still suffer from sample efficiency and require significant amounts of training data.

	\subsubsection{Physics-based prediction models}
    Several works have investigated differentiable formulations of physics engines which could be embedded as layers in deep neural networks. 
	In~\cite{degrave2019_diffphys} an impulse-based velocity stepping physics engine is implemented in a deep learning framework.
	Collisions are restricted to sphere shapes and sphere-plane interactions to allow for automatic differentiation.
	The method is used to tune a deep-learning based robot controller but neither demonstrated for parameter identification nor video prediction.
	
	    Belbute-Peres et al.~\cite{belbuteperes2018_diffphys} propose an end-to-end differentiable physics engine that models frictions and collisions between arbitrary shapes.
    Gradients are computed analytically at the solution of the resulting linear complementarity problem (LCP)~\cite{optnet}. 
    They demonstrate the method for including a differentiable physics layer in a video prediction network for modelling a 2D bouncing balls scenario with 3 color-coded circular objects.
    Input to the network are the color segmented images and optical flow estimated from pairs of frames.
    The network is trained in a supervised way using ground-truth positions of the objects.
    We propose to use spatial transformers in the decoder such that the network can learn a video representation in a self-supervised way.
    We investigate 3D scenarios that include pushing, sliding, and collisions of objects and analyze observability of physical parameters using vision and known forces applied to the objects.
		A different way of formulating rigid body dynamics has been investigated in~\cite{wallach2019_hamnn} using energy conservation laws. 
	The method is demonstrated for parameter identification, angle estimation and video prediction for a 2D pendulum environment using an autoencoder network.
	Similar to our approach,~\cite{jaques2020_physasinvgraph} also uses spatial transformers for the decoder.
	However, differently the physics engine only models gravitational forces between objects and does not investigate full 3D rigid body physics with collision and friction modelling and parameter identification.
	
	Recently, Runia et al.~\cite{runia2020_clothinthewind} demonstrated an approach for estimating physical parameters of deforming cloth in real-world scenes.
	The approach minimizes distance in a contrastively learned embedding space which encodes videos of the observed scene and rendered scenes generated with a physical model based on the estimated parameters.
	In our approach, we train a video embedding network with the physical model as network layer and identify the physical parameters of observed rigid objects during training.

	\section{Background}
	\label{sec:background}
	\subsection{Unconstrained and Constrained Dynamics}
	The governing equation of unconstrained rigid body dynamics in 3D can be written as 
	\begin{equation}
		\mathbf{f} = \mathbf{M}\dot{\boldsymbol{\xi}} + \textrm{Coriolis forces}
		\label{eq: short newton euler}
	\end{equation}
	where $\mathbf{f}: [0,\infty[ ~ \rightarrow ~ \mathbb{R}^{6}$ is the time-dependent torque-force vector, $\mathbf{M} \in \mathbb{R}^{6 \times 6}$ is the mass-inertia matrix and $\dot{\boldsymbol{\xi}}: [0,\infty[ ~ \rightarrow ~ \mathbb{R}^6$ is the time-derivative of the twist vector  $\boldsymbol{\xi} = \left( \boldsymbol{\omega}^\top, \mathbf{v}^\top \right)^\top$ stacking rotational and linear velocities $\boldsymbol{\omega}, \mathbf{v}: [0,\infty[ \rightarrow \mathbb{R}^3$~\cite{Cline_2002}. 
	In our experiments we do not consider rotations between two or more frames of reference, therefore we do not have any Coriolis forces. 
	Most of the real world rigid body motions are constrained. 
	To simulate those behaviors we need to constrain the motion with joint, contact and frictional constraints \cite{Cline_2002}.
	
	The force-acceleration based dynamics which we use in equation \eqref{eq: short newton euler} does not work well for collisions since there is a sudden change in the direction of velocity in infinitesimal time \cite{Cline_2002}. Therefore we use impulse-velocity based dynamics, where even the friction is well-behaved \cite{Cline_2002}, i.e., equations have a solution at all configurations. We discretize the acceleration using forward Euler method as $\dot{\boldsymbol{\xi}} = (\boldsymbol{\xi}_{t+h} - \boldsymbol{\xi}_t)/h$, where $\dot{\boldsymbol{\xi}}_{t+h}$ and $\dot{\boldsymbol{\xi}}_t$ are the velocities in successive time steps at times $t+h$ and $t$, and $h$ is the time-step size. Equation \eqref{eq: short newton euler} now becomes
	\begin{equation}
		\mathbf{M}\boldsymbol{\xi}_{t+h} = \mathbf{M}\boldsymbol{\xi}_{t} + \mathbf{f} \cdot h.
		\label{eq: discretized newton euler}
	\end{equation}
	\subsubsection{Constrained Dynamics:} The joint constraints are equality constraints and they restrict degrees of freedom of a rigid body. Mathematically this can be written as $\mathbf{J}_e\boldsymbol{\xi}_{t+h}=0$ where $\mathbf{J}_e$ is the equality Jacobian which gives the directions in which the motion is restricted. The joint constraints exert constraint forces which are solved using Euler-Lagrange equations by solving for the joint force multiplier $\boldsymbol{\lambda}_e$.
	
	The contact constraints are inequality constraints which prevent bodies from interpenetration. This ensures that the minimum distance between two bodies is always greater than or equal to zero. The constraint equations can be written using Newton's impact model \cite{Cline_2002} as $\mathbf{J}_c\boldsymbol{\xi}_{t+h} \geq -k\mathbf{J}_c\boldsymbol{\xi}_{t}$. The term $k\mathbf{J}_c\boldsymbol{\xi}_{t}$ can be replaced with $\mathbf{c}$ which gives $\mathbf{J}_c\boldsymbol{\xi}_{t+h} \geq -\mathbf{c}$, where $k$ is the coefficient of restitution, $\mathbf{J}_c$ is the Jacobian of the contact constraint function at the current state of the system and $\boldsymbol{\lambda}_c$ is the contact force multiplier. Since it is an inequality constraint we introduce slack variables $\mathbf{a}$, which also gives us complementarity constraints \cite{boyd}.
	
	The friction is modeled using a maximum dissipation energy principle since friction damps the energy of the system. In this case we get two inequality constraints since frictional force depends on normal force \cite{FrictionModeling,FrictionModeling2}. They can be written as $\mathbf{J}_f\boldsymbol{\lambda}_f + \boldsymbol{\gamma} \geq 0$ and $\mu \boldsymbol{\lambda}_c \geq \mathbf{E}\boldsymbol{\lambda}_f$ where $\mu$ is the friction coefficient, $\mathbf{J}_f$ is the Jacobian of the friction constraint function at the current state of the system, $\mathbf{E}$ is a binary matrix which ensures linear independence between equations at multiple contacts, and $\boldsymbol{\lambda}_f$ and $\boldsymbol{\gamma}$ are frictional force multipliers. Since we have two inequality constraints we have two slack variables $\boldsymbol{\sigma},\boldsymbol{\zeta}$ and two complementarity constraints.
	
	In summary, all the constraints that describe the dynamic behavior of the objects we consider in our scene can be written as the following linear complementarity problem (LCP),
    
	\begin{equation}
		\begin{pmatrix}
			0\\
			0\\
			\mathbf{a}\\
			\boldsymbol{\sigma}\\
			\boldsymbol{\zeta}
		\end{pmatrix}
		-
		\begin{pmatrix}
			\mathbf{M} && -\mathbf{J}_{eq}^T && -\mathbf{J}_{c}^T && -\mathbf{J}_{f}^T && 0\\
			\mathbf{J}_{eq} && 0 && 0 && 0 && 0\\
			\mathbf{J}_{c} && 0 && 0 && 0 && 0\\
			\mathbf{J}_{f} && 0 && 0 && 0 && \mathbf{E}\\
			0 && 0 && \mu && -\mathbf{E}^T && 0
		\end{pmatrix}
		\begin{pmatrix}
			\boldsymbol{\xi}_{t+h}\\
			\boldsymbol{\lambda}_{eq}\\
			\boldsymbol{\lambda}_{c}\\
			\boldsymbol{\lambda}_{f}\\
			\boldsymbol{\gamma}
		\end{pmatrix}
		=
		\begin{pmatrix}
			-\mathbf{M}\boldsymbol{\xi}_{t} - h\mathbf{f}_{ext}\\
			0\\
			\mathbf{c}\\
			0\\
			0
		\end{pmatrix},\\
	\label{eq: friction matrix dynamics impulse}
	\end{equation}
	\begin{equation*}
		\textrm{subject to}\:\:\begin{pmatrix}
			\mathbf{a}\\
			\boldsymbol{\sigma}\\
			\boldsymbol{\zeta}
		\end{pmatrix}\geq 0,
		\:\:\begin{pmatrix}
			\boldsymbol{\lambda}_c\\
			\boldsymbol{\lambda}_f\\
			\boldsymbol{\gamma}
		\end{pmatrix}\geq 0,
		\:\:\begin{pmatrix}
			\mathbf{a}\\
			\boldsymbol{\sigma}\\
			\boldsymbol{\zeta}
		\end{pmatrix}^T\begin{pmatrix}
			\boldsymbol{\lambda}_c\\
			\boldsymbol{\lambda}_f\\
			\boldsymbol{\gamma}
		\end{pmatrix}=0.
	\end{equation*}

	The above LCP is solved using a primal-dual algorithm as described in~\cite{boyd}.
	It is embedded in our deep neural network architecture in a similar way as in \cite{optnet} and \cite{belbuteperes2018_diffphys}, which facilitates backpropagation of gradients at its solution.

	\section{Method}
	\label{sec:method}

    We develop a deep neural network architecture which encodes images into physical states $\mathbf{s}_i = \left( \mathbf{x}_i^\top, \boldsymbol{\xi}^\top \right)^\top$ where $\mathbf{x}_i = \left( \mathbf{q}_i^\top, \mathbf{p}_i^\top \right)^\top$ with orientation $\mathbf{q}_i \in \mathbb{S}^3$ as unit quaternion and position $\mathbf{p}_i \in \mathbb{R}^3$ of object $i$. 
    We propagate the state using the differential physics engine which is integrated as layer on the encoding in the deep neural network.
    For self-supervised learning, a differentiable decoder subnetwork generates images from the integrated state representation of the objects.

	We aim to learn the system's dynamics by regressing the state trajectories and learning the physical parameters of the objects. 
	These parameters can be the masses of the bodies and the coefficient of friction between two bodies. We initialize the objects at certain locations in the scene with some velocity and start the simulation by applying forces. 
	In the following, we will detail our network architecture and training losses.

	\subsection{Network Architecture}
	\label{sec:architecture}

    \subsubsection{Encoder}
    For supervised learning experiments, we use convolutional layers followed by fully connected layers with exponential linear units (ELU) \cite{elu} to encode poses from images.
    The encoder receives the image $I_t$ and is encoded as pose $\textbf{x}_t$. 
	We need at least two images to infer velocities from images. 
	For self-supervised learning experiments, we use a variational encoder~\cite{vae} with the same base architecture as in the supervised case.
	Here, the encoder receives the image $I_t$ and outputs the mean and log variance of a Gaussian hidden state distribution $p( \mathbf{z}_t \mid I_t )$. 
	A latent space sample is obtained using the reparameterization trick~\cite{vae}, and the pose $\textbf{x}_t$ is regressed from it using a fully connected layer.
	
	We use three images so that we can average out the velocities in case of collisions when the two frames are collected just before and after collision. 
	We use the difference in poses to estimate velocity instead of directly training the network to output velocities. 
	This gives us the average velocity, not the final velocity. For example in 1D, when a block of mass $m$ is acting under an effective force $f_{\text{eff}}$ between times $t_0$ and $t_1$, the velocity at time $t_1$ is given by
	\begin{equation}
        v(t_1) = \underbrace{\frac{p(t_1) - p(t_0)}{t_1 - t_0}}_{\text{average\;velocity}} + \frac{1}{2}\frac{f_{\text{eff}}}{m}(t_1 - t_0)
        \label{eq: final velocity from poses}
	\end{equation}
    If we would let the network learn the velocities, it would require to implicitly learn the physics which we want to avoid by the use of the differentiable physics engine.
	The encoded states are provided as input to the differentiable physics engine.

    \subsubsection{Trajectory Integration}
	We integrate a trajectory of poses from the initial pose estimated by the encoder and the velocity estimates by the differentiable physics engine.
	In each time step, we calculate the new pose of each object $\mathbf{x} = \left( \mathbf{q}^\top, \mathbf{p}^\top \right)^\top$ where $\mathbf{q} \in \mathbb{S}^3$ is a unit quaternion representing rotation and $\mathbf{p} \in \mathbb{R}^3$ is the position from the resulting velocities of the LCP $\boldsymbol{\xi}_t = \left(\boldsymbol{\omega}_t^\top, \mathbf{v}_t^\top \right)^\top$ by
	\begin{equation}
	    \begin{split}
    		\mathbf{p}_t &= \mathbf{p}_t + \mathbf{v}_t\cdot h\\
    		\mathbf{q}_{t} &= \mathbf{q}_{t} \times \textrm{quat}(e^{0.5\boldsymbol{\omega}_t h})
		\end{split}
	\end{equation}
	where $\textrm{quat}(\cdot)$ is an operator which converts a rotation matrix into a quaternion.

    \subsubsection{Decoder}
    We use a spatial transformer network layer~\cite{stn} to decode object poses into images. 
    These transformations provide structure to the latent space and thus allow the network to train in a self-supervised way. 
    The poses estimated by the physics engine or inferred by the encoder are given as inputs to the network along with content patches of the objects and the background. 
    The content patches and the background are assumed known and extracted from training images using ground-truth masks.
    The spatial transformer renders these objects at appropriate locations on the background assuming the camera intrinsics, its view pose in the scene and parameterization of the plane are known.

    \subsection{Training Losses}

	\subsubsection{Supervised Learning}
	\label{sec:suploss}
    For supervised learning, we initialize the physics engine with inferred poses $\mathbf{x}^{enc}_{1:N,i}$ for each object $i$ from the encoder where $N$ is the (video) sequence length. 
    Estimated poses $\hat{\mathbf{x}}_{1:N,i}$ by the physics engine as well as the inferred poses by the encoder are compared with ground truth poses $\mathbf{x}^{gt}_{1:N,i}$ to infer physical parameters,
	\begin{equation}
        \label{eq: Encoder loss}
	    \begin{split}
            L_{\textrm{supervised}} &=  \sum_i e( \mathbf{x}^{gt}_{1:N,i}, \mathbf{x}^{enc}_{1:N,i} ) + \alpha e( \mathbf{x}^{gt}_{1:N,i}, \hat{\mathbf{x}}_{1:N,i} ),\\
            e( \mathbf{x}_1, \mathbf{x}_2 ) &:= \frac{1}{N} \sum_{t=1}^N \left\| \ln( \mathbf{q}_{2,t}^{-1} \mathbf{q}_{1,t} ) \right\|_2^2 + \left\| \mathbf{p}_{2,t} - \mathbf{p}_{1,t} \right\|_2^2,\\ 
        \end{split}
	\end{equation}	
	where $\alpha$ is a weighting constant $\alpha=0.1$ in our experiments), $t$ is the time step and $i$ indexes objects.
	We use the quaternion geodesic norm to measure differences in rotations.
	
	\subsubsection{Self-supervised Learning}
	\label{sec:selfsuploss}
    For self-supervised learning, we initialize the physics engine with inferred poses $\mathbf{x}^{enc}_{1:N}$ from the encoder.
    Both estimated poses by the physics engine and the inferred poses are reconstructed into images $\hat{I}^{rec}_{1:N}$ and $I^{rec}_{1:N}$ using our decoder, respectively.
    The images are compared to input frames $I^{gt}_{1:N}$ to identify the physical parameters and train the network.
    We impose a KL divergence loss between the inferred encoder distribution $p( \mathbf{z}_t \mid I_{t} )$ and the standard normal distribution prior $p( \mathbf{z}_t ) = \mathcal{N}( \mathbf{0}, \mathbf{I} )$,
	\begin{multline}
        L_{\textrm{self-supervised}} = \frac{1}{N} \left\| I^{gt}_{1:N} - I^{rec}_{1:N} \right\|_2^2 + \frac{\alpha}{N} \left\| I^{gt}_{1:N} - \hat{I}^{rec}_{1:N} \right\|_2^2\\ + \sum_{t=1}^N \mathit{KL}\left( p( \mathbf{z}_t \mid I_t ) \| p( \mathbf{z}_t ) \right).
        \label{eq: Encoder decoder loss}
	\end{multline}

	\subsubsection{System Identification}
	\label{sec:sysid}
	For reference, we also directly optimize for the physical parameters based on the ground-truth trajectories $\mathbf{p}^{gt}_{1:N}$ without the image encoder.
	For this we use the first state as an input to the differentiable physics engine.
	In this case, the loss function is
	
       $L_{\textrm{sys-id}} = \sum_i e( \mathbf{x}^{gt}_{i}, \hat{\mathbf{x}}_{i} ).$

    \section{Experiments}
	\label{sec:experiments}
	
\begin{figure}[tb]
    \centering
            \includegraphics[trim={0 0 0 2cm},clip,width=0.19\textwidth]{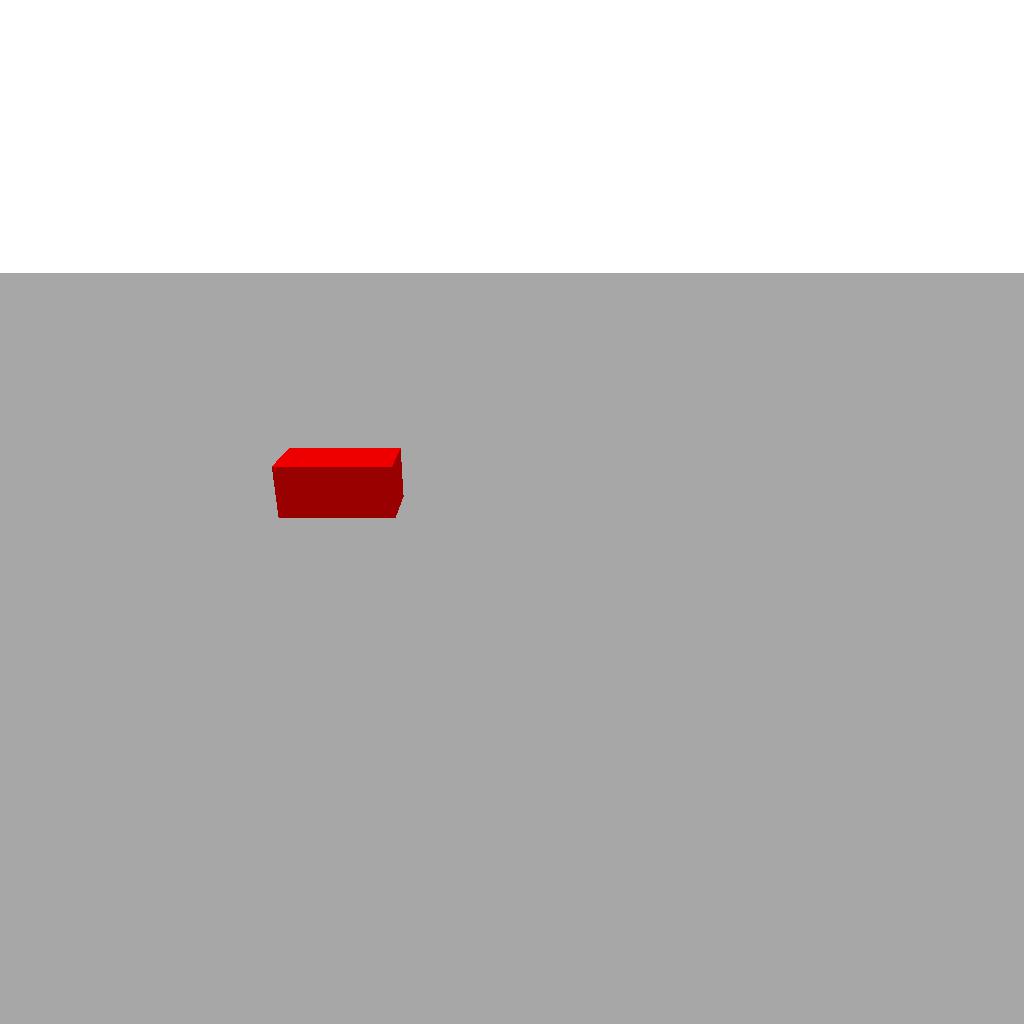}
            \includegraphics[trim={0 0 0 2cm},clip,width=0.19\textwidth]{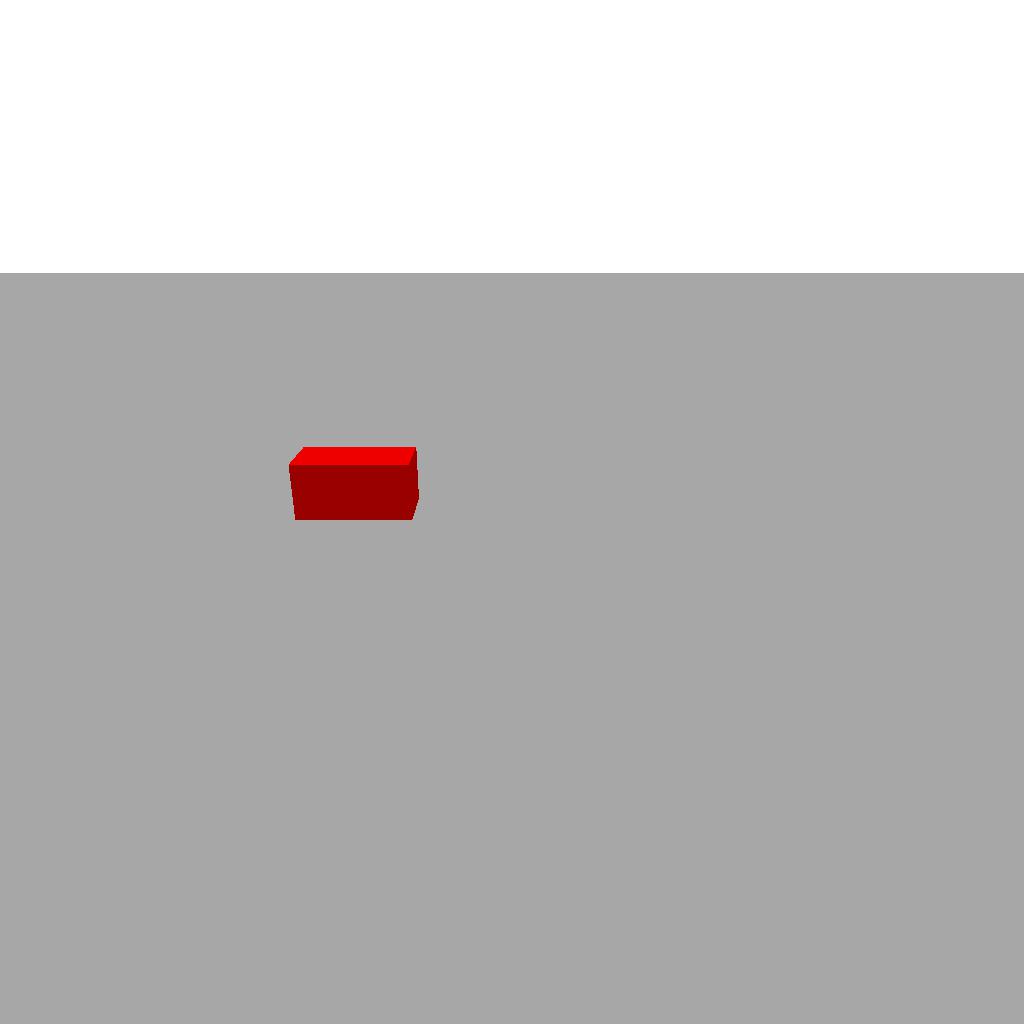}
            \includegraphics[trim={0 0 0 2cm},clip,width=0.19\textwidth]{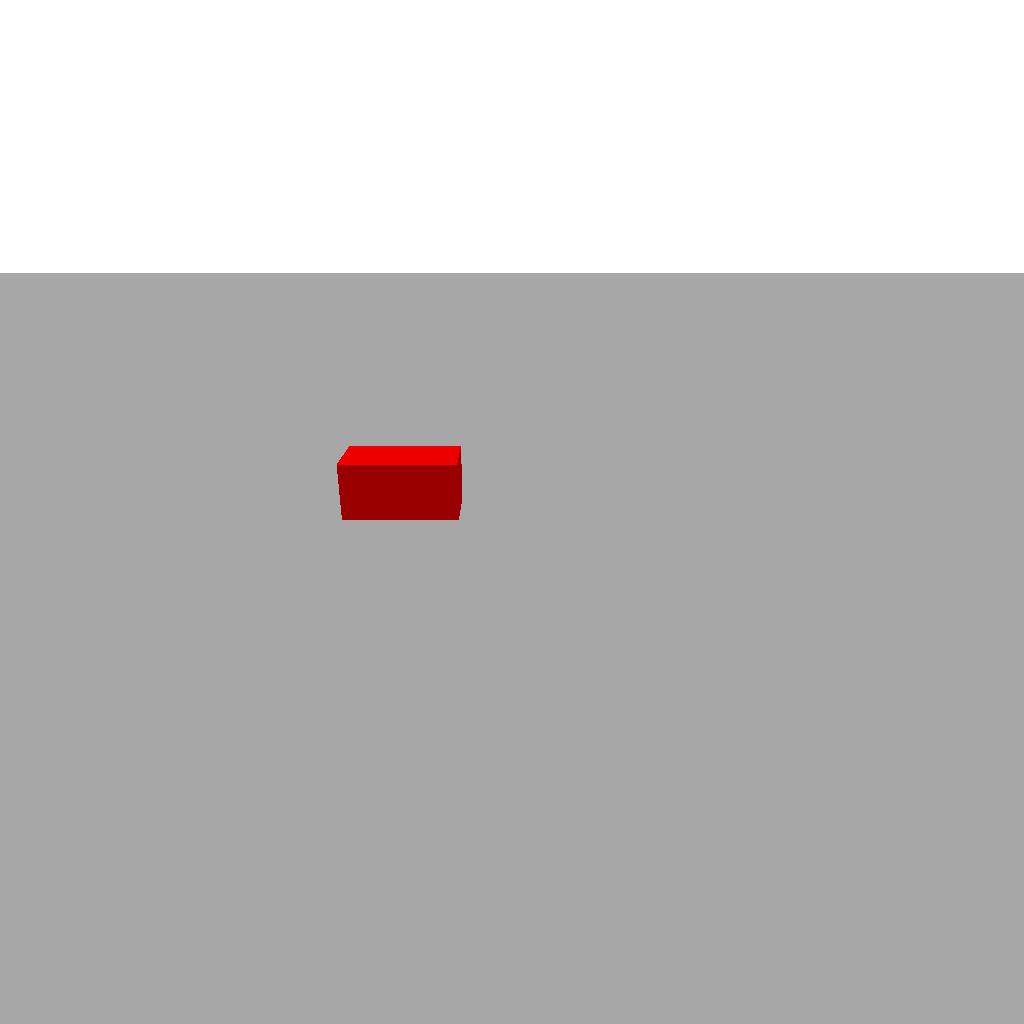}
            \includegraphics[trim={0 0 0 2cm},clip,width=0.19\textwidth]{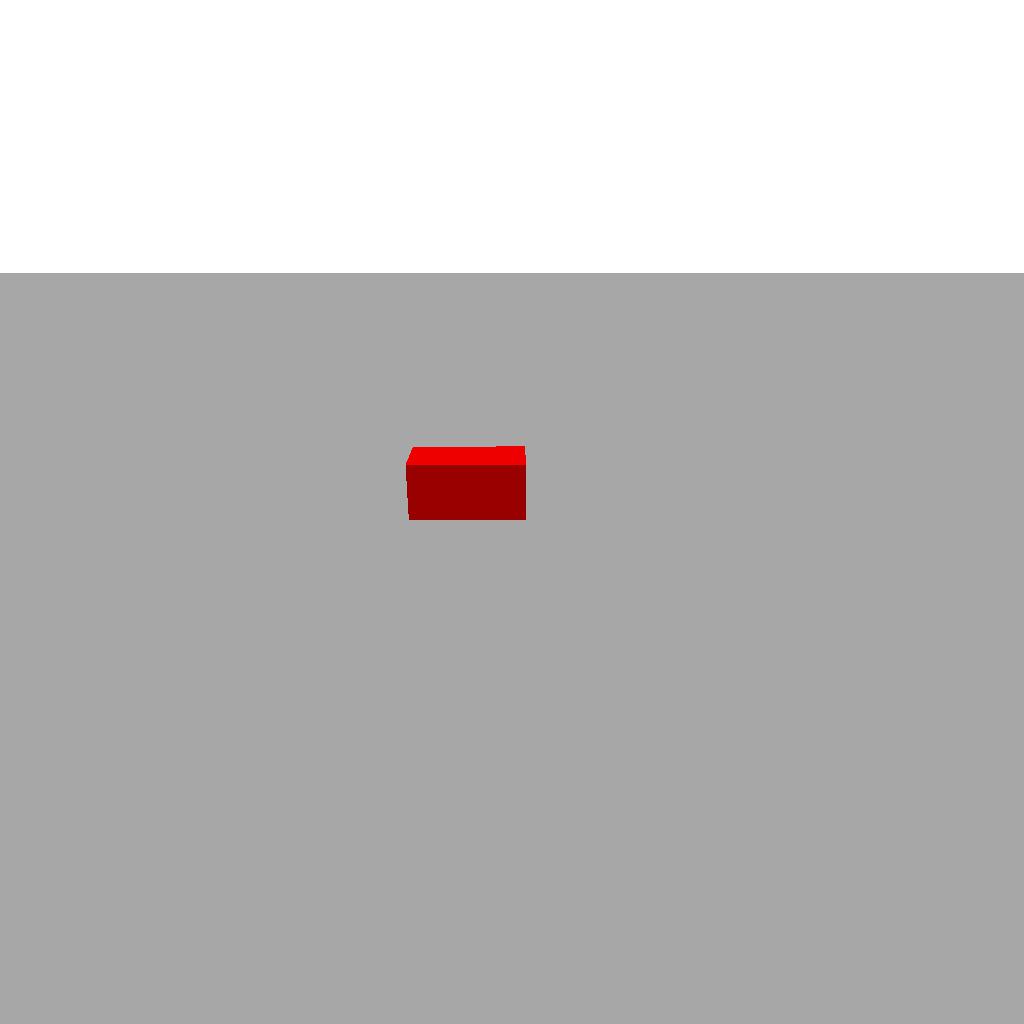}
            \includegraphics[trim={0 0 0 2cm},clip,width=0.19\textwidth]{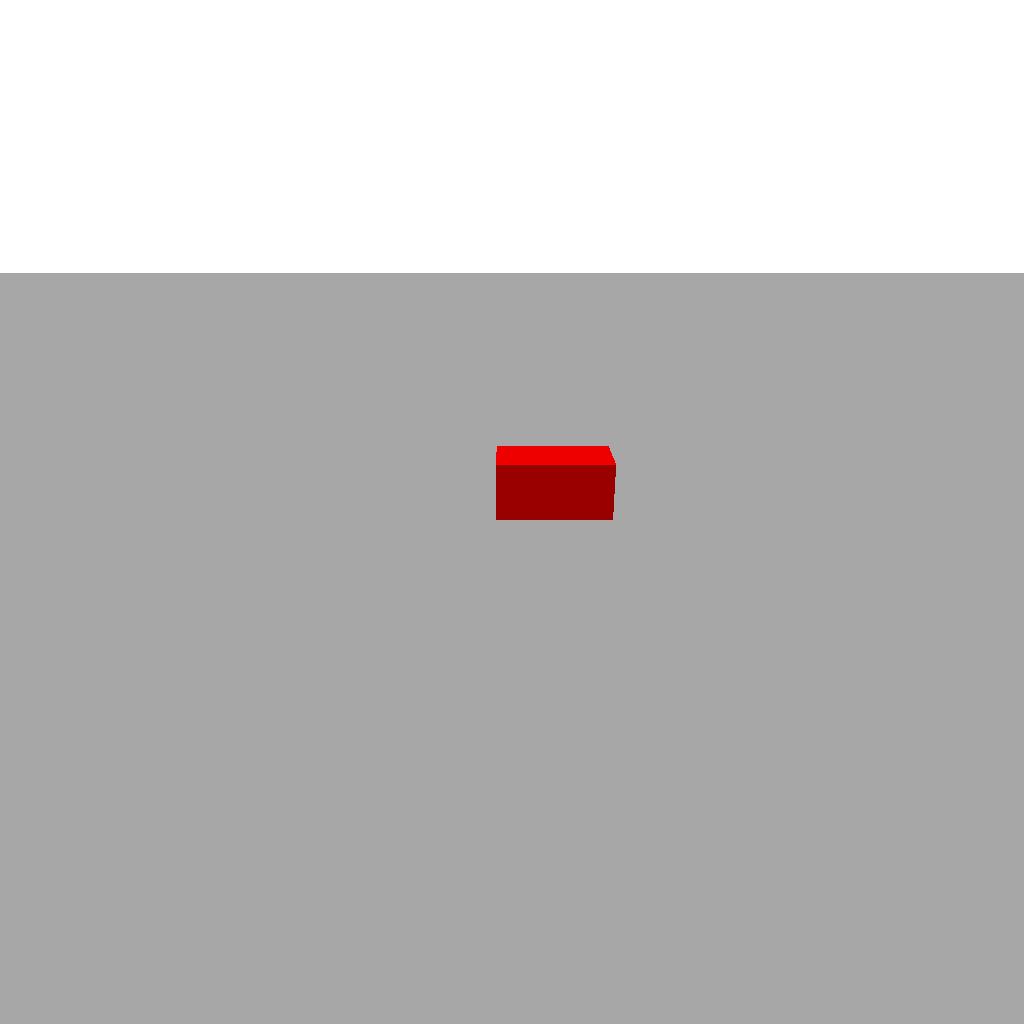}\\
            \includegraphics[trim={0 0 0 8cm},clip,width=0.19\textwidth]{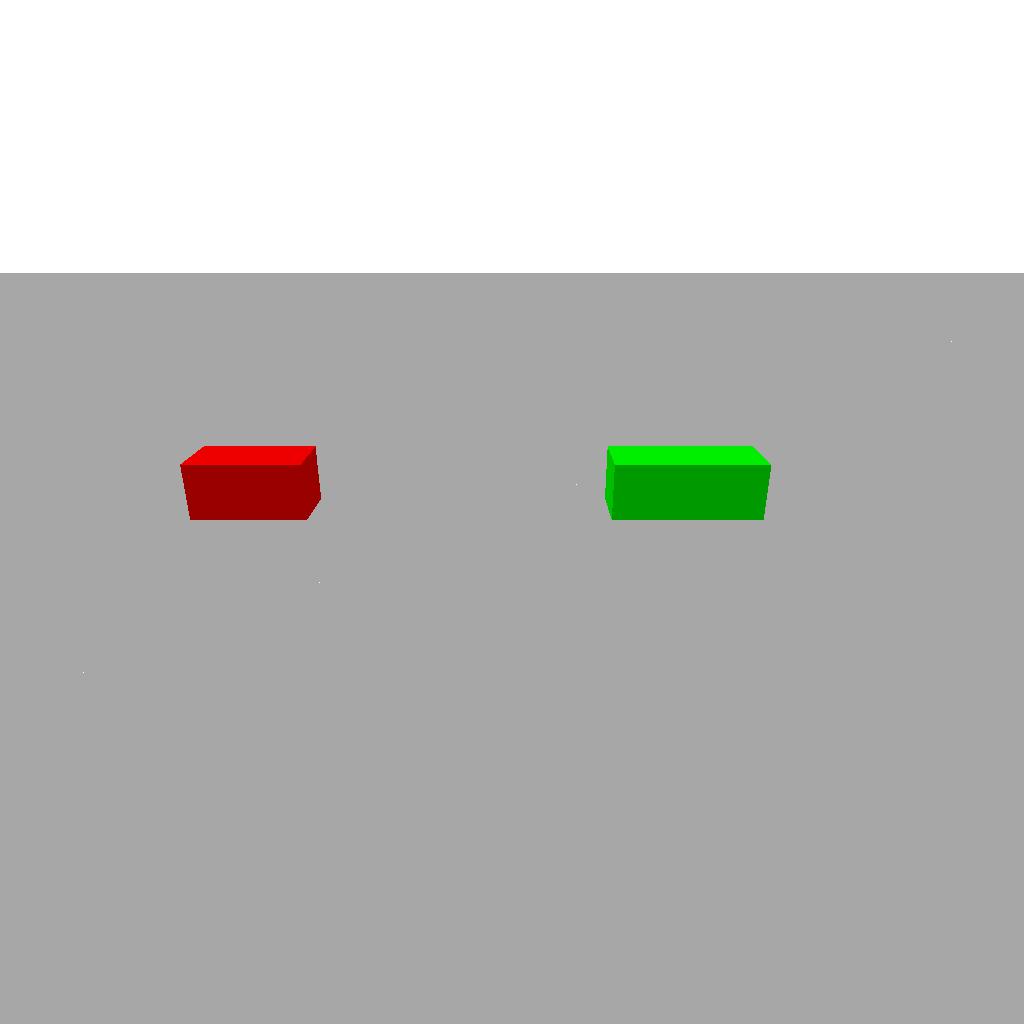}
            \includegraphics[trim={0 0 0 8cm},clip,width=0.19\textwidth]{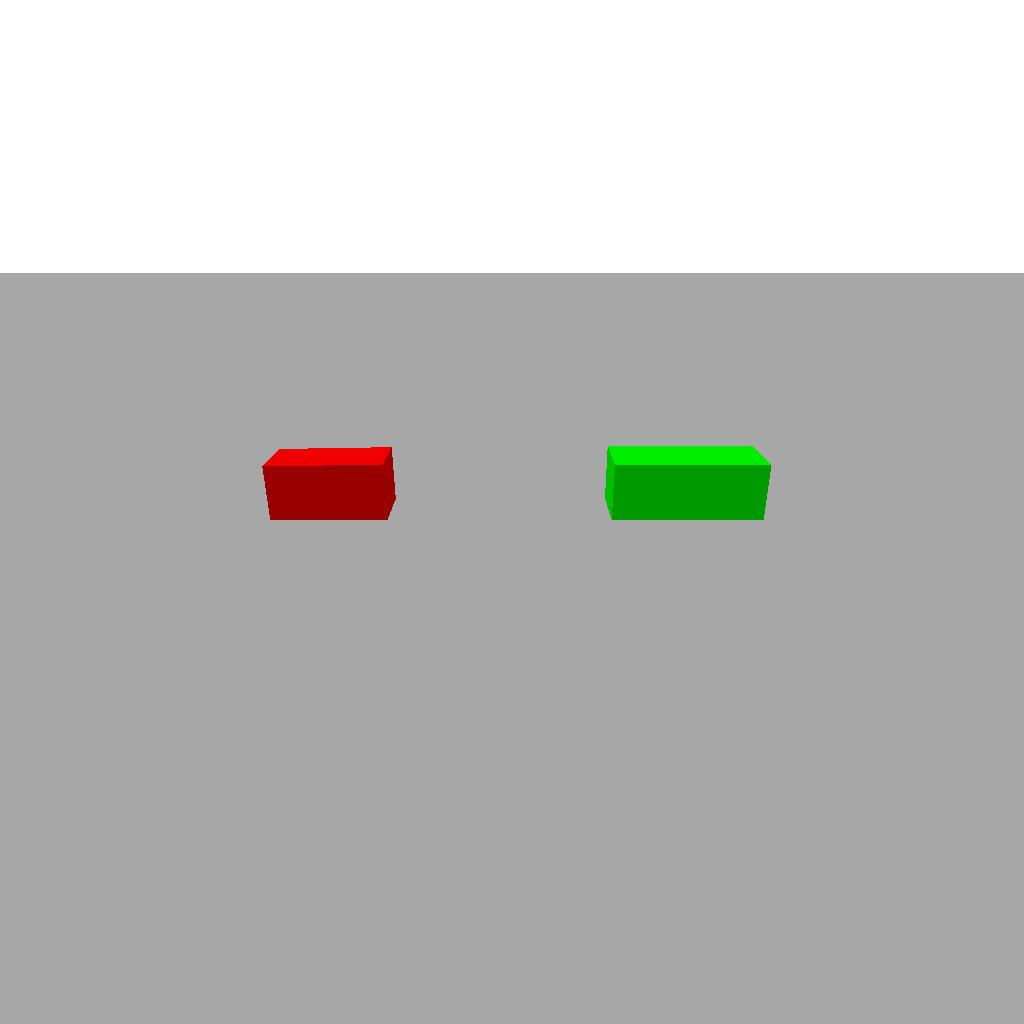}
            \includegraphics[trim={0 0 0 8cm},clip,width=0.19\textwidth]{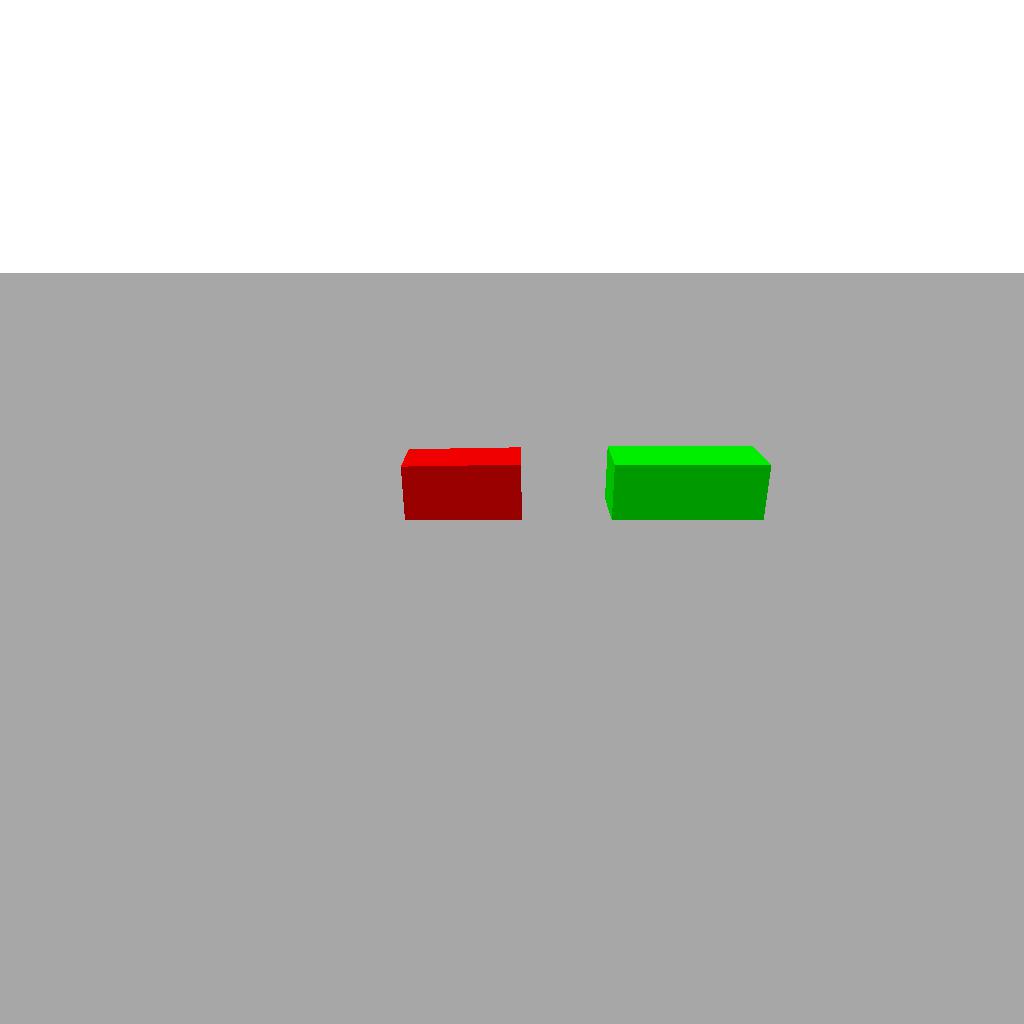}
            \includegraphics[trim={0 0 0 8cm},clip,width=0.19\textwidth]{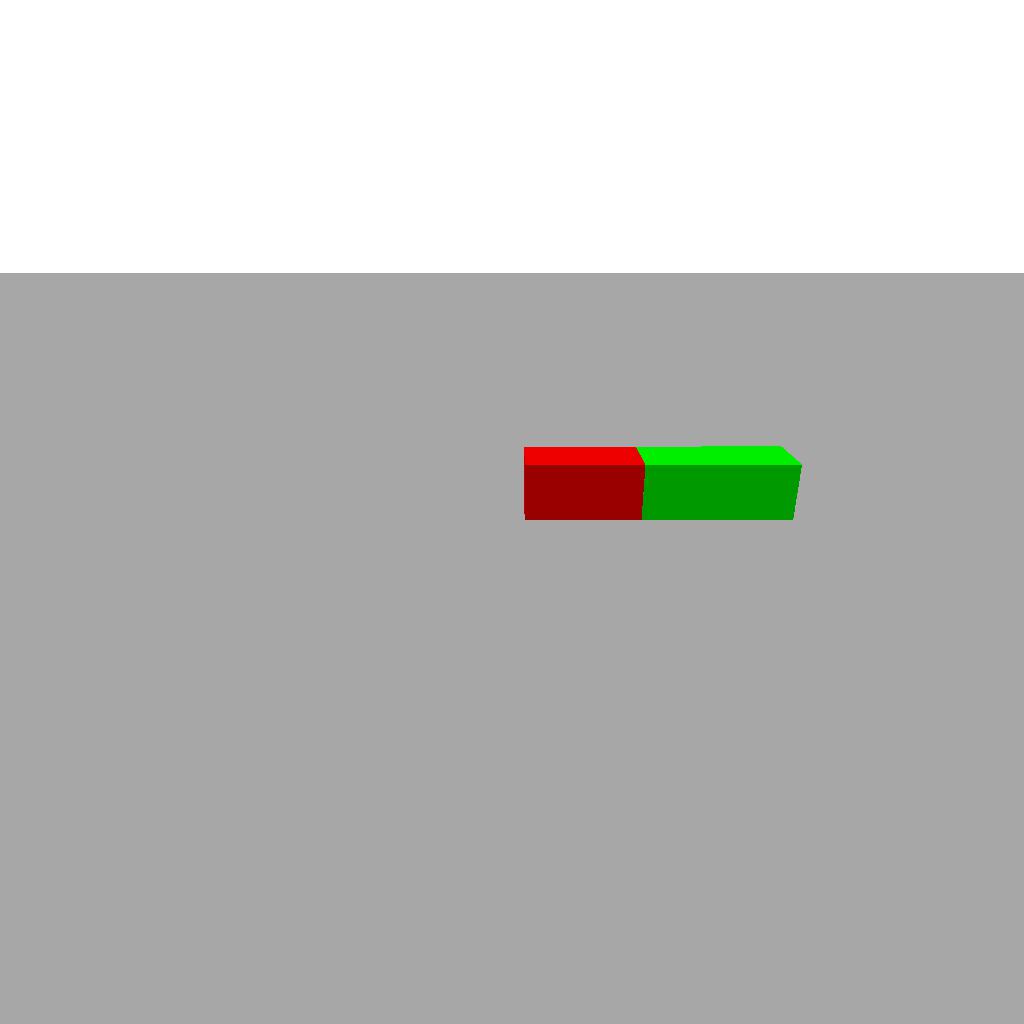}
            \includegraphics[trim={0 0 0 8cm},clip,width=0.19\textwidth]{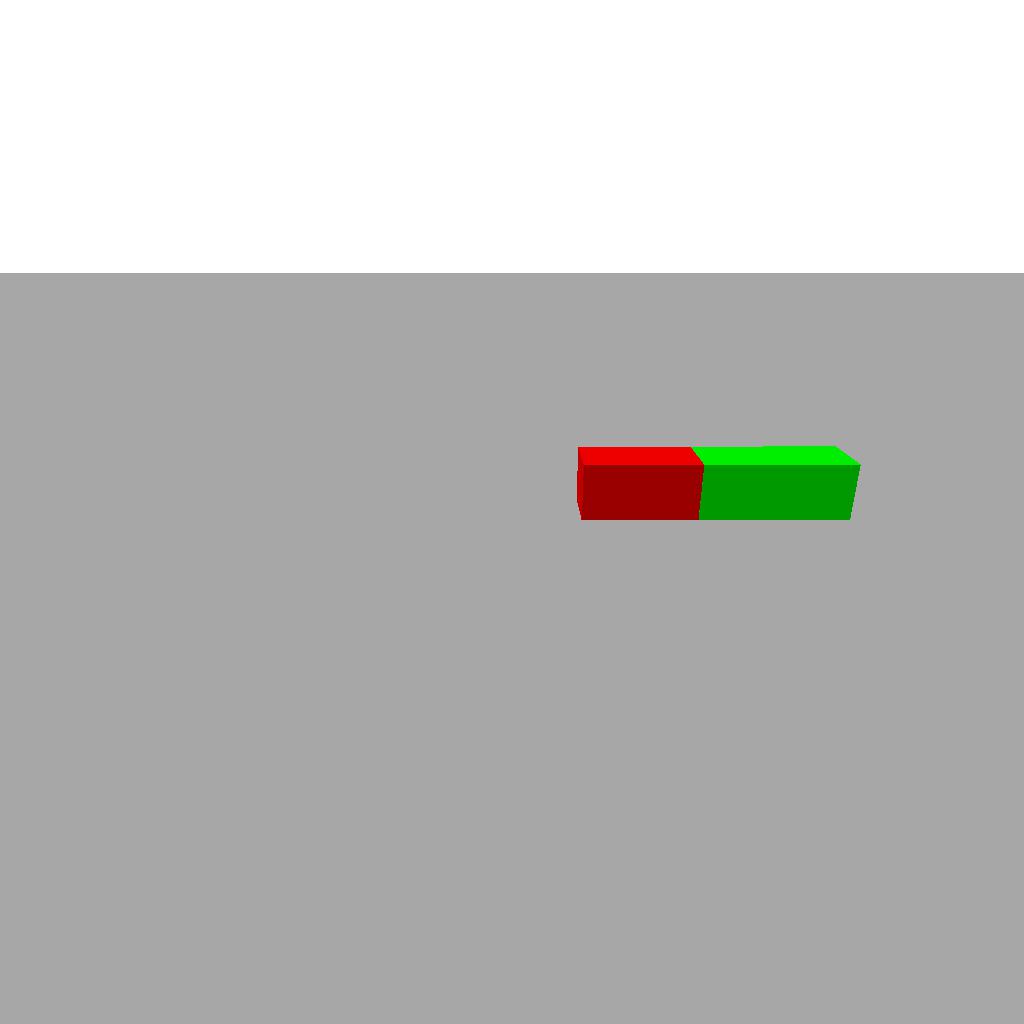}\\
            \includegraphics[trim={0 0 0 4cm},clip,width=0.19\textwidth]{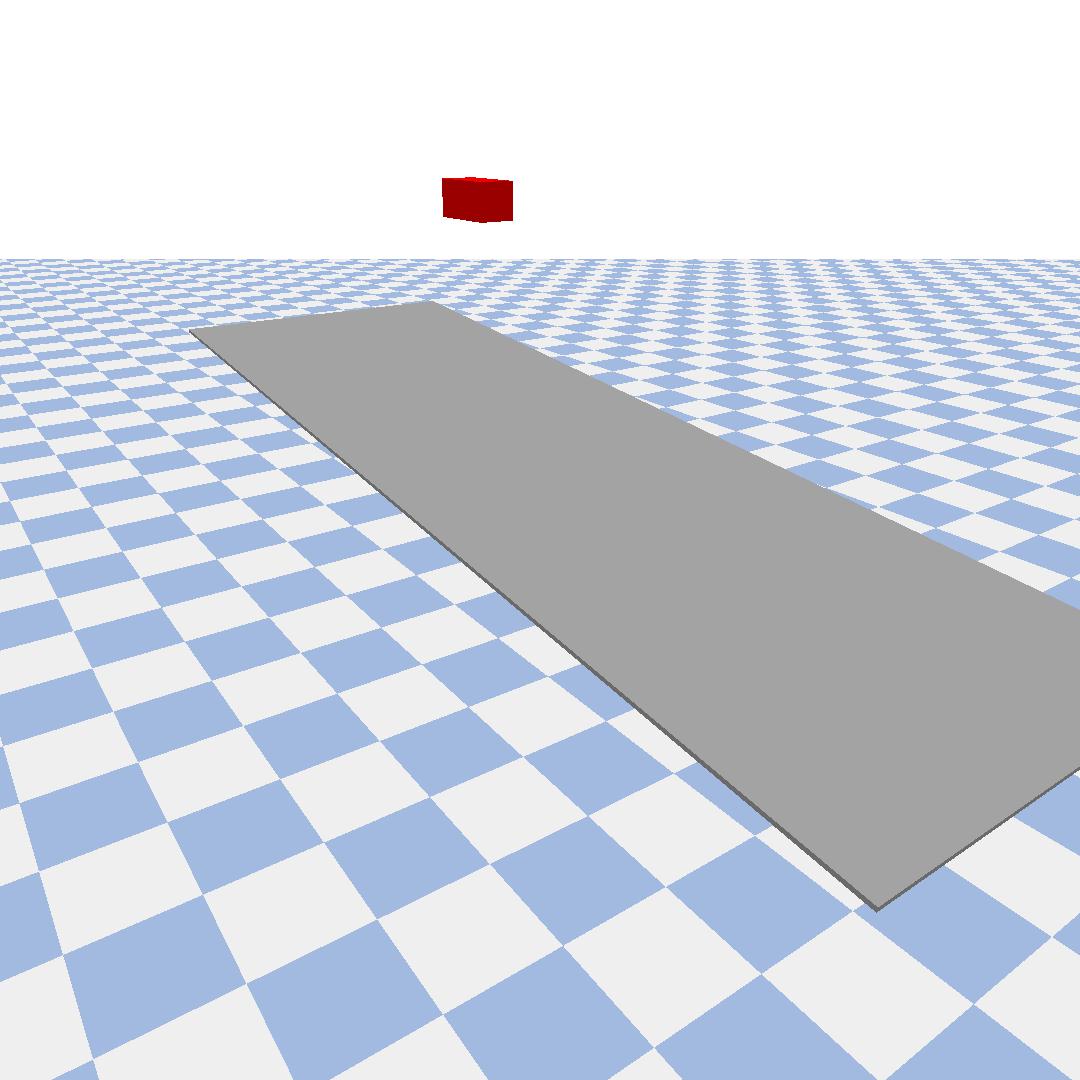}
            \includegraphics[trim={0 0 0 4cm},clip,width=0.19\textwidth]{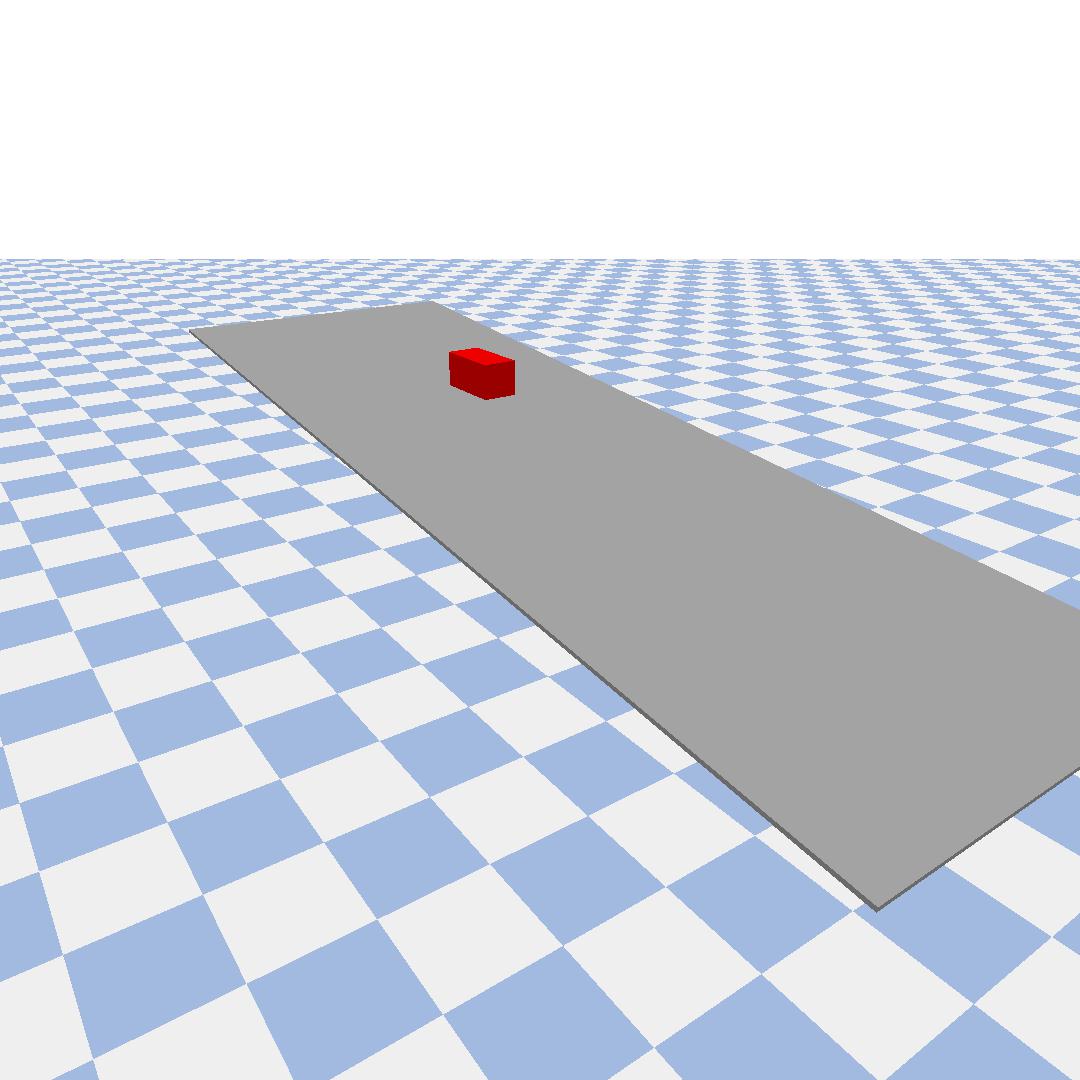}
            \includegraphics[trim={0 0 0 4cm},clip,width=0.19\textwidth]{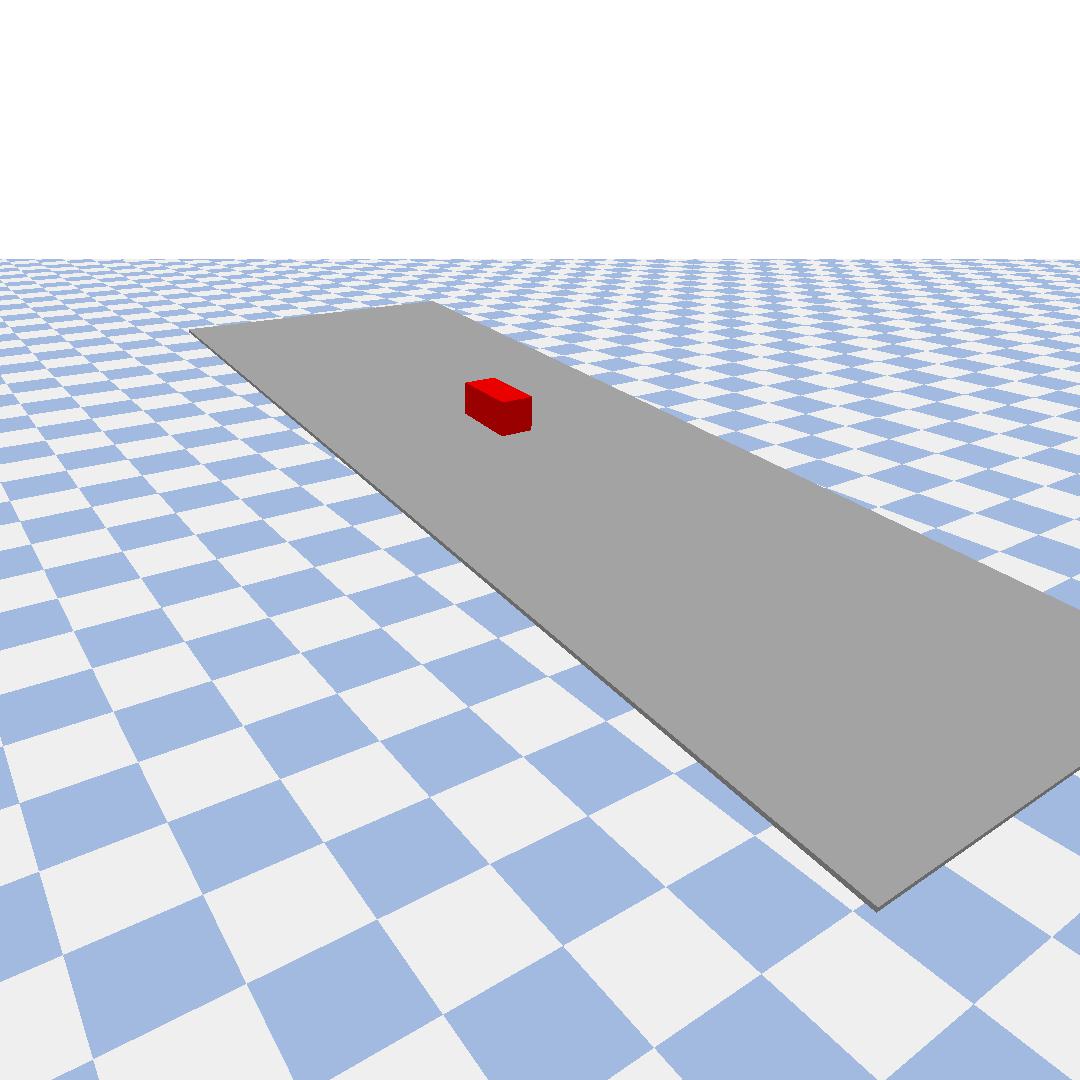}
            \includegraphics[trim={0 0 0 4cm},clip,width=0.19\textwidth]{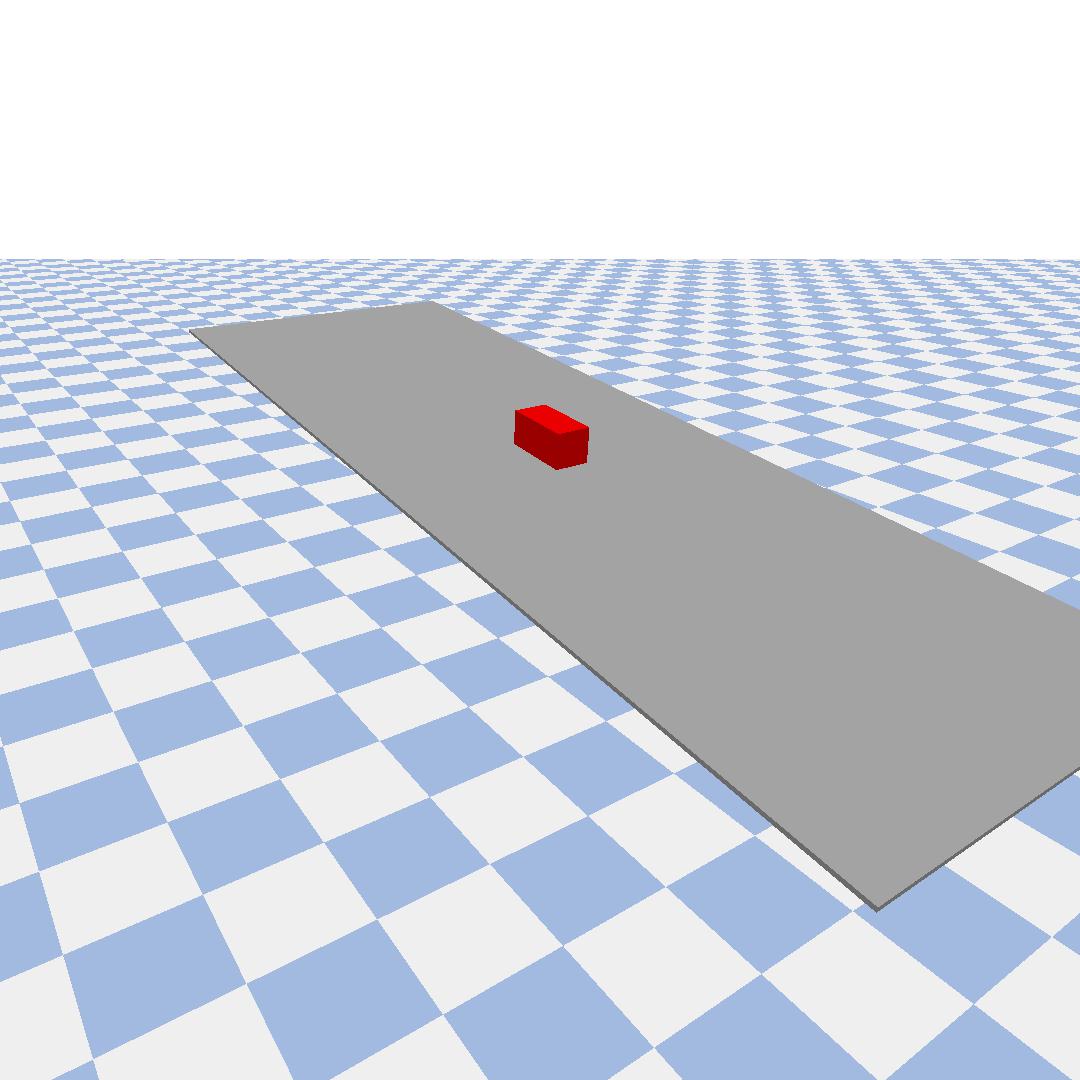}
            \includegraphics[trim={0 0 0 4cm},clip,width=0.19\textwidth]{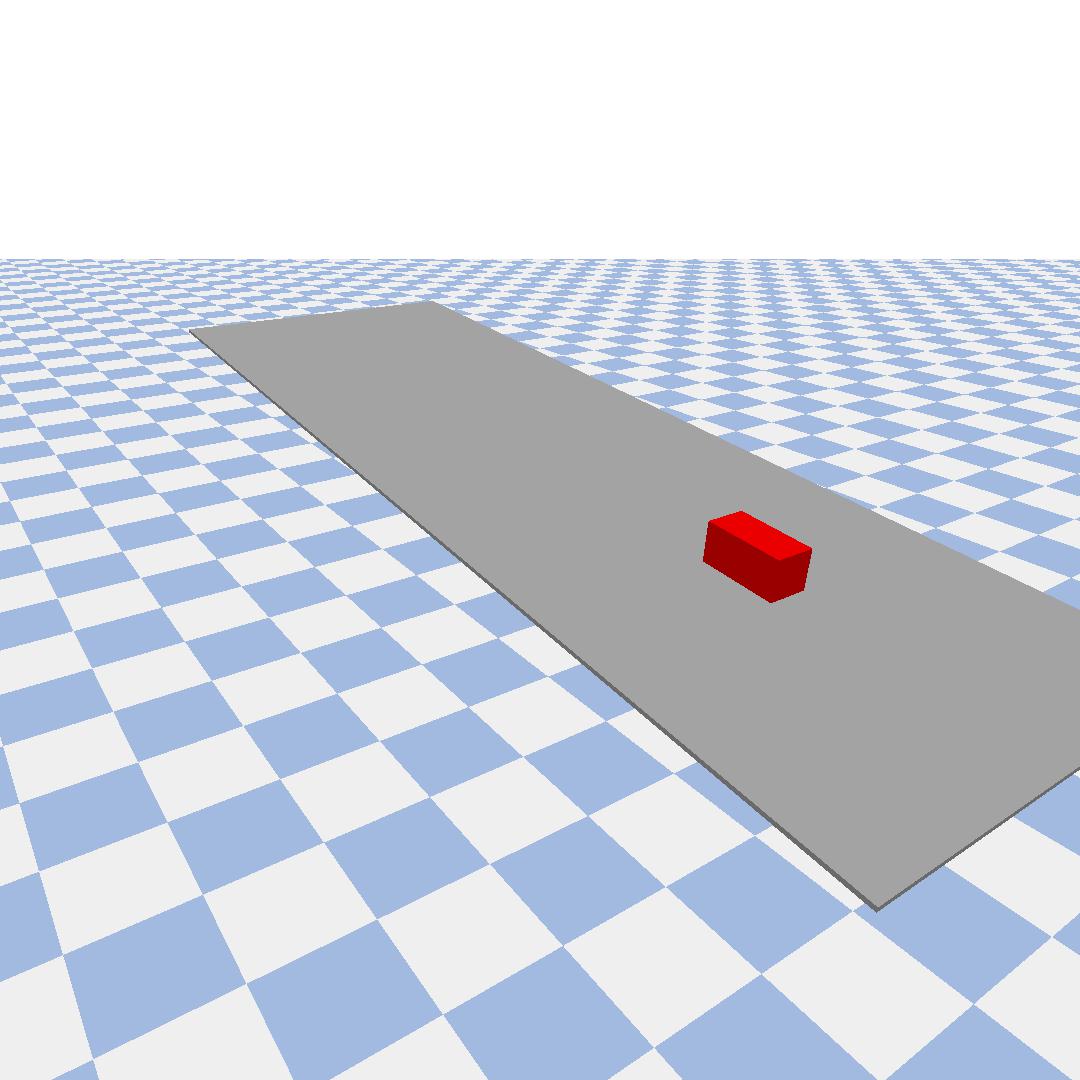}
        \caption{3D visualization of the simulated scenes. Top: block pushed on a flat plane. Middle: block colliding with another block. Bottom: block falling and sliding down on an inclined plane.}
        \label{fig:3dscenes}
	\end{figure}

	We evaluate our approach in 3D simulated scenarios including pushing, sliding and collision of objects (see Fig.~\ref{fig:3dscenes}).
	
	\subsection{Simulated Scenarios and Observability Analysis}
	In this section, we discuss and analyze the different scenarios for the observability of physical parameters.
	To this end, we simplify the scenarios into 1D or 2D scenarios where dynamics equations are simpler to write.

	\begin{figure}[h!]
        \centering
        \begin{minipage}[c]{0.49\textwidth}
        \centering
        \includegraphics[scale=0.4]{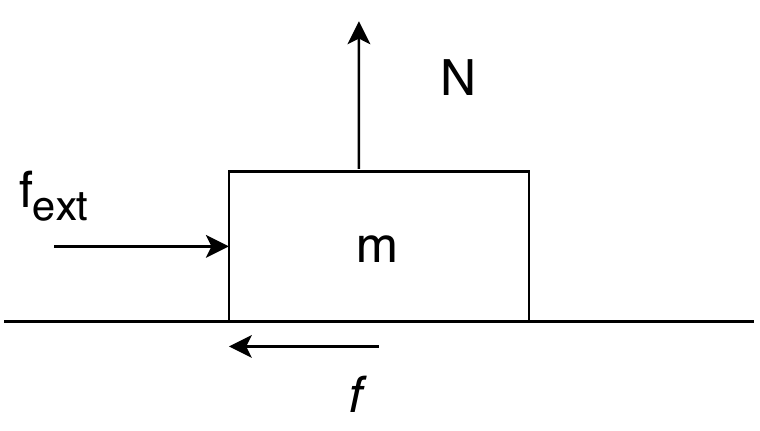}\\\vspace{2ex}
        \includegraphics[scale=0.4]{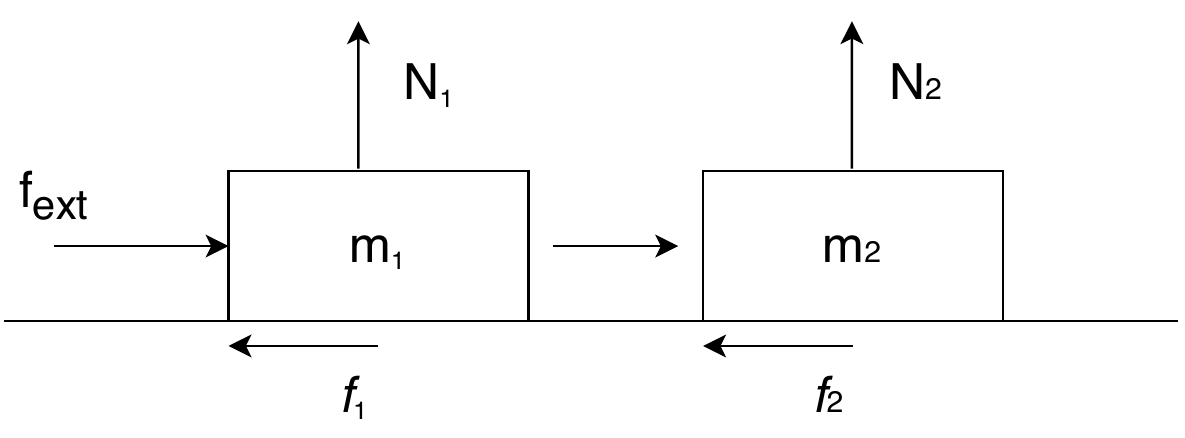}
        \end{minipage}
        \begin{minipage}[c]{0.49\textwidth}
        \centering
        \includegraphics[scale=0.4]{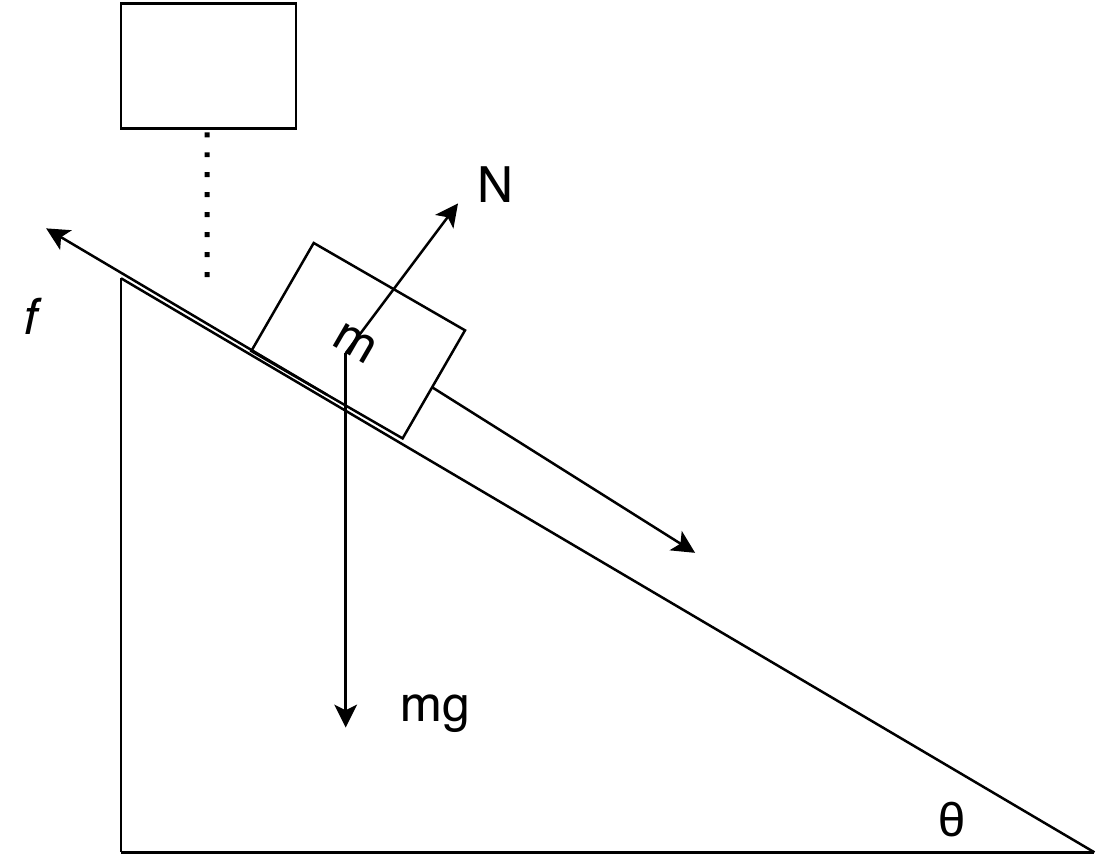}
        \end{minipage}
        \caption{1D/2D sketches of scenarios. Top left: block pushed on a flat plane. Bottom left: block colliding with another block. Right: block sliding down on an inclined plane.}
        \label{fig:1ddiagrams}
	\end{figure}

    \subsubsection{Block Pushed On a Flat Plane}
	In this scenario, a block of mass $m$, lying on a flat plane is pushed with a force $\textbf{f}_{ext}$ at the center of mass as shown in Fig.~\ref{fig:1ddiagrams} (top left). 
	In this 1D example, since we only have a frictional constraint we can use Eq.~\eqref{eq: discretized newton euler} in combination with the frictional force $f = \mu N$ to describe the system, where $\mu$ is the coefficient of friction, $g=9.81m/s^2$ is the acceleration due to gravity and $N=mg$ is the normal force since the body has no vertical motion.
	The velocity $v_{t+h}$ in the next time step hence is
	\begin{equation}
        v_{t+h} = v_t + \frac{f_{ext}}{m}h - \mu gh
        \label{eq: velocity update}
	\end{equation}
    We observe that only either one of mass or friction can be inferred at a time. 
    Thus, in our experiments we fix one of the parameters and learn the other.

	\subsubsection{Block Colliding With Another Block} 
	To learn both mass and coefficient of friction simultaneously, we introduce a second block with known mass ($m_2$) made of the same material like the first one. 
	This ensures that the coefficient of friction ($\mu$) between the plane and the two blocks is same. 
	Since we are pushing the blocks, after collision, both blocks move together. 
	In the 1D example in Fig.~\ref{fig:1ddiagrams} (bottom left), when applied an external force ($f_{\text{ext}}$), the equation to calculate the linear velocities $v_{1/2,t+h}$ of both objects in the next time step becomes
	\begin{equation}
            v_{1,{t+h}} = v_{1_t} + \frac{f_{\text{ext}}}{m_1}h - \mu gh,~~~ v_{2,{t+h}} = v_{2_t} + \frac{f'}{m_2}h - \mu gh,
	\end{equation}
	where $\mu g m_1$ and $\mu g m_2$ are frictional forces acting on each block and $f'$ is the equivalent force on the second body when moving together. 
	Now, in our experiments we can learn both mass and coefficient of friction together given the rest of the parameters in the equation.
    
    \subsubsection{Block Freefall and Sliding Down On an Inclined Plane}
	In this scenario the block slides down the inclined plane after experiencing a freefall as shown in Fig.~\ref{fig:1ddiagrams} (right). 
	In the 1D example, since the freefall is unconstrained (ignoring air resistance), the velocity update is given by $v_{t+h} = v_t + gh$. 
	For block sliding down on an inclined plane, the equation to calculate velocity in the next time is
		$v_{t+h} = v_t + g(\sin{\theta} - \mu \cos{\theta})\;h$,
	where $\theta$ is the plane inclination.
	We can see that we can only infer the coefficient of friction $\mu$ and due to the free fall we do not need to apply additional forces.

	\subsection{Results}
	\label{results}
	We simulated the scenarios in 3D using the bullet physics engine using PyBullet~\footnote{https://pybullet.org}.
	Note that the bullet physics engine is different to the LCP physics engine in our network and can yield qualitatively and numerically different results.
	The bodies are initialized at random locations to cover the whole workspace.
	Random forces between $1$--$10$\,N are applied at each time step. 
	These forces are applied in $+x$, $-x$, $+y$ and $-y$ directions which are chosen at random but kept constant for a single trajectory while the magnitude of the forces randomly varies in each time step. 
	In total, 1000 different trajectories are created with 300 time steps each for each scenario. 
	We render top-down views at $128 \times 128$ resolution. 
	Training and test data are split with ratio~$9:1$. 
	For evaluation we show the evolution of the physical parameters during the training.
	We also give the average relative position error by the encoder which is the average of the difference between ground truth positions and estimated poses divided by object size. 

	\begin{table}[tb]
        \centering
        \setlength{\tabcolsep}{0.5em}
        \begin{tabular}{m{1.5cm} m{3.1cm} m{3.1cm} m{3.1cm}}
        \toprule
            Inference & Block Pushed On a Flat Plane & Block Sliding Down the Inclined Plane & Block Colliding With Another Block\\
            \midrule
            Mass &
            \includegraphics[scale=0.065]{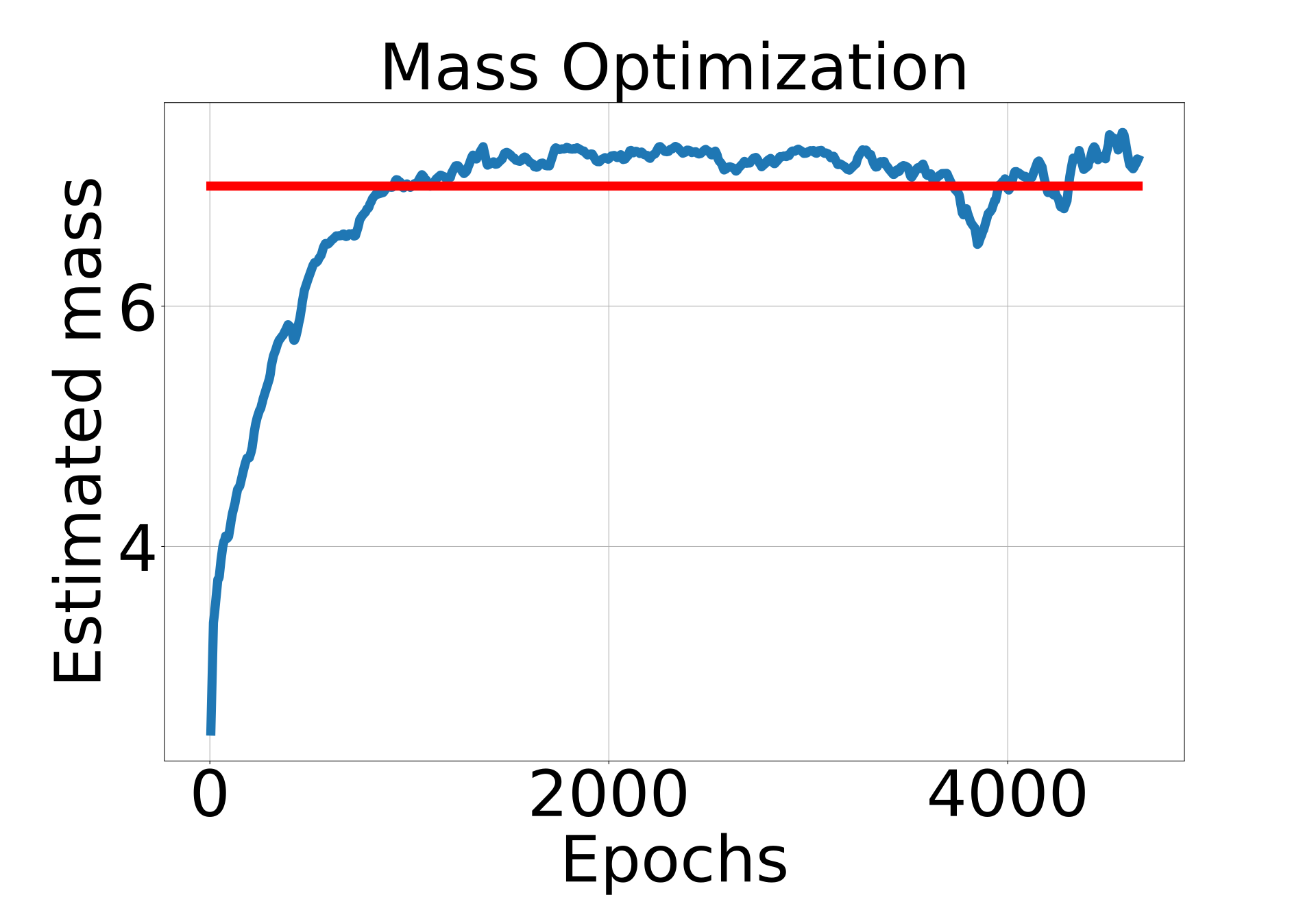} \tiny{position inference error: 4\%}
            &
            \centering Not feasible
            & 
            \includegraphics[scale=0.065]{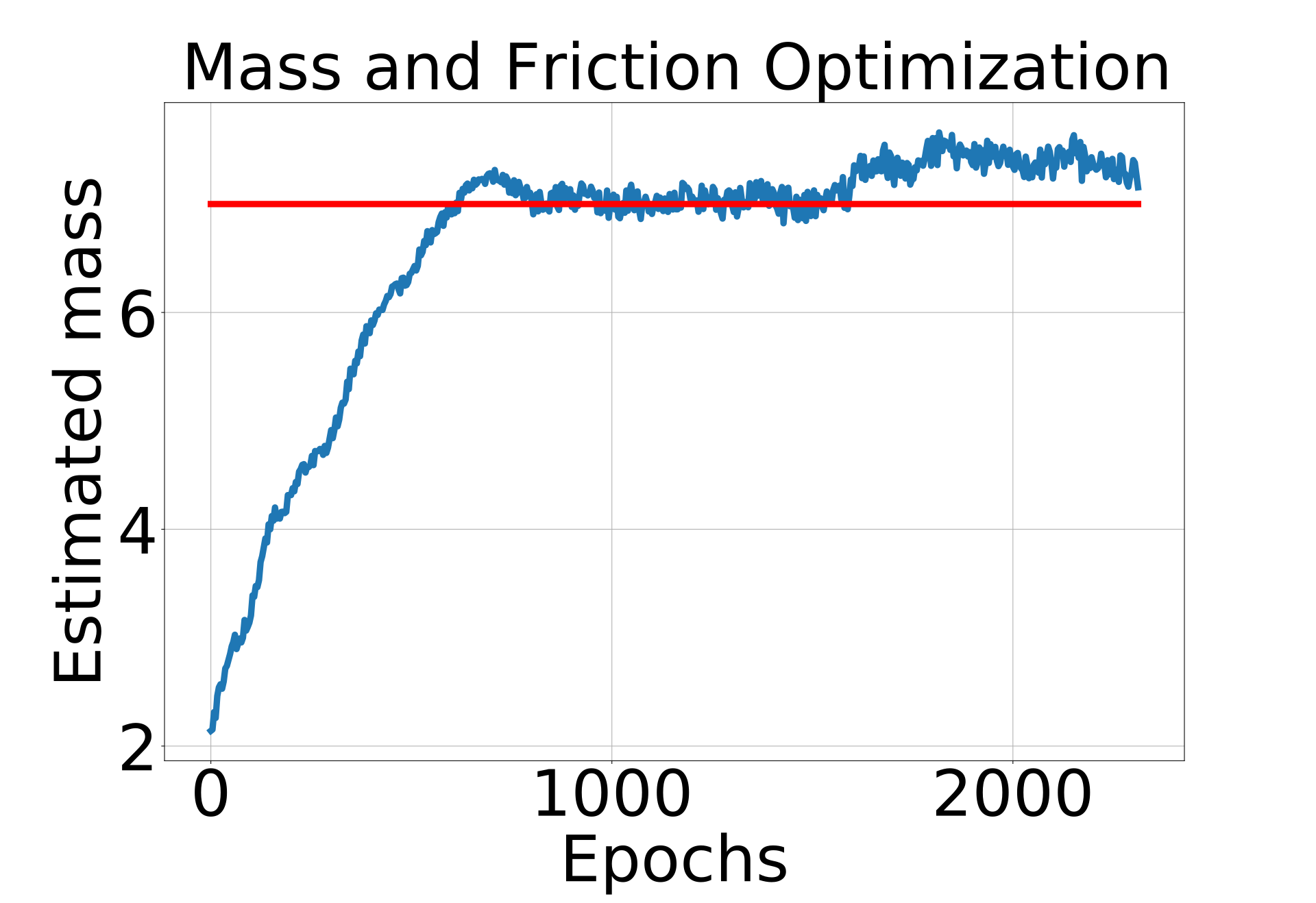} \tiny{position inference error: 8\%}\\
            \midrule
            Coefficient of friction &
            \includegraphics[scale=0.065]{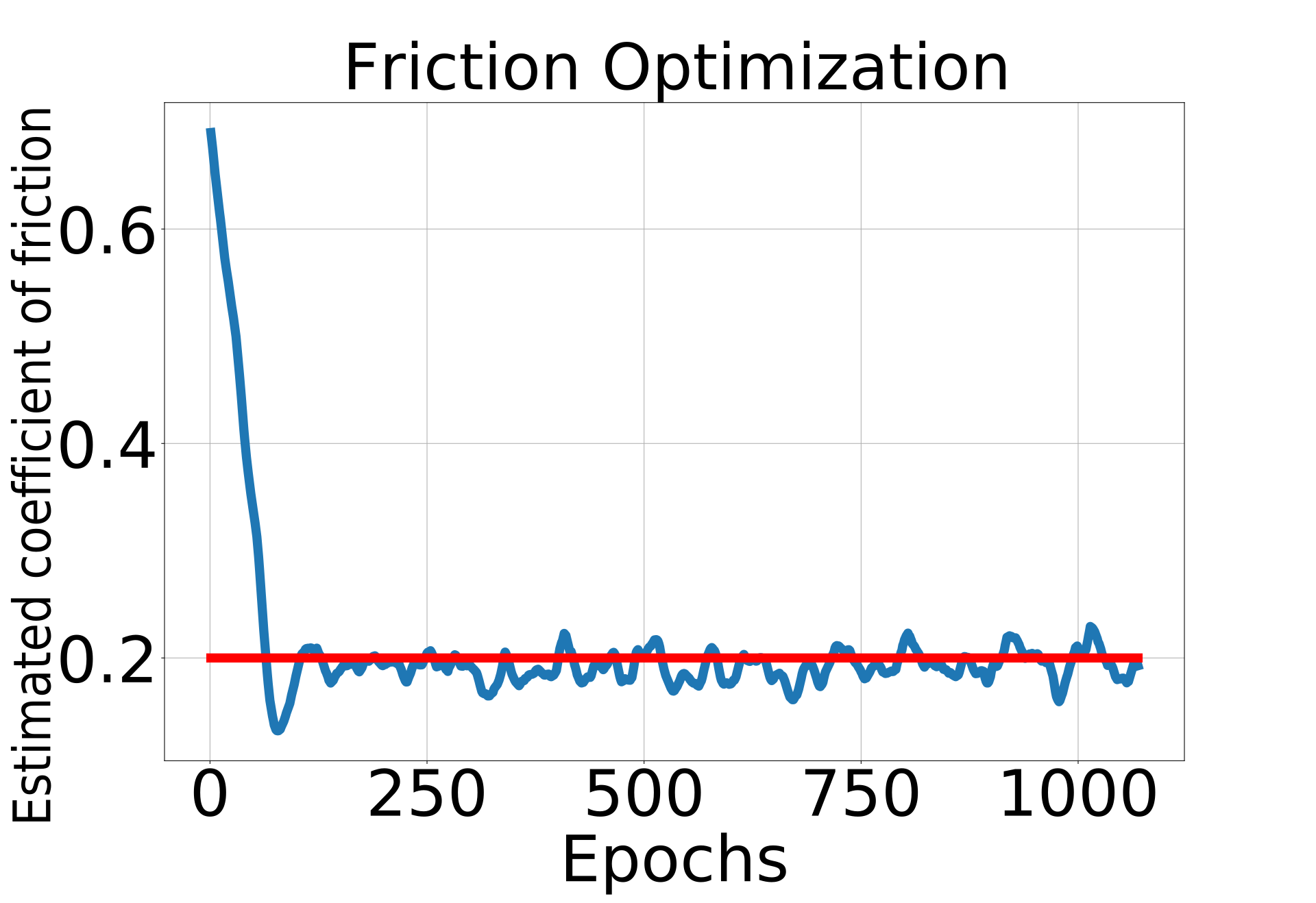} \tiny{position inference error: 2\%}
            &
            \includegraphics[scale=0.065]{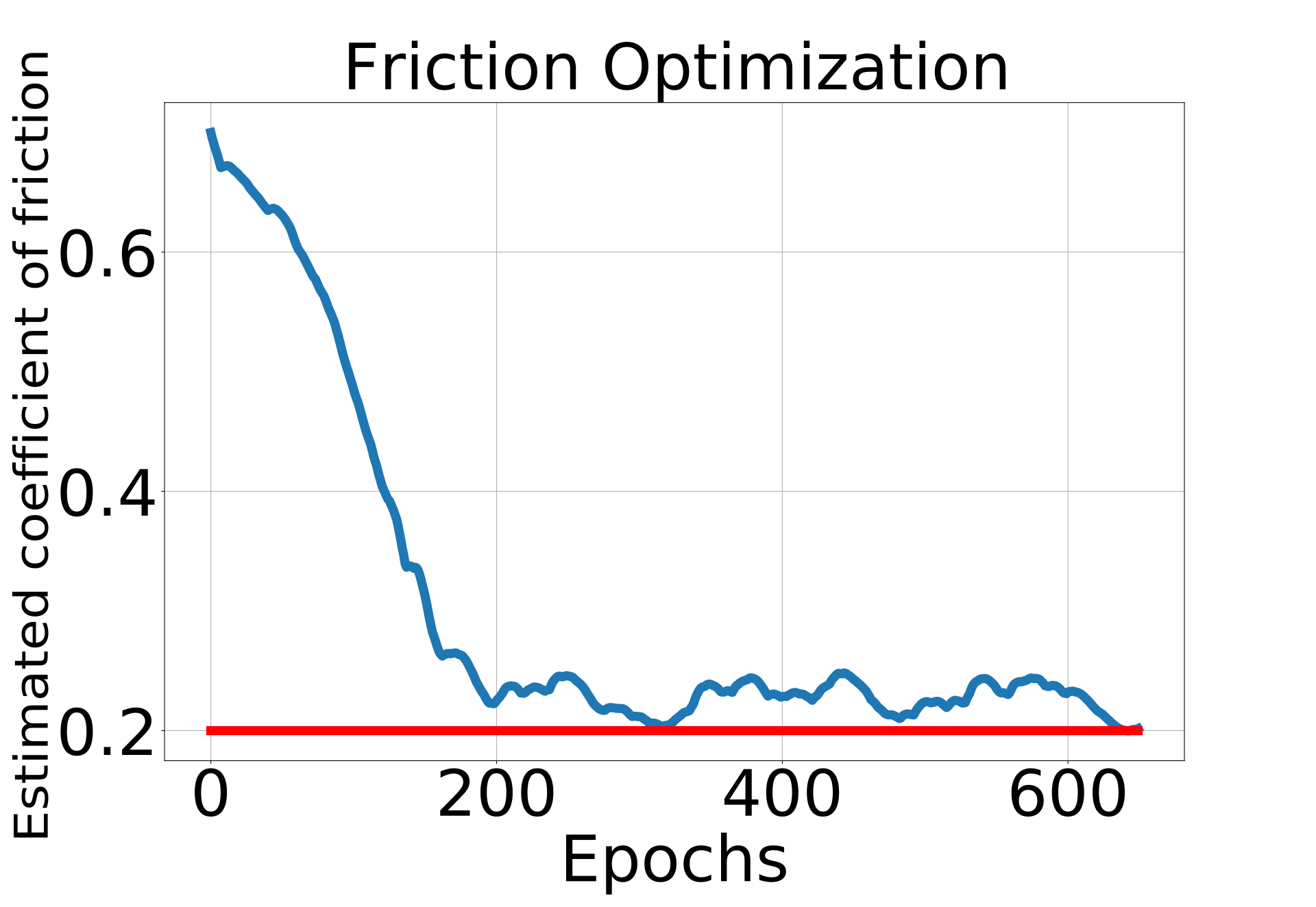} \tiny{position inference error: 5\%}
            & 
            \includegraphics[scale=0.065]{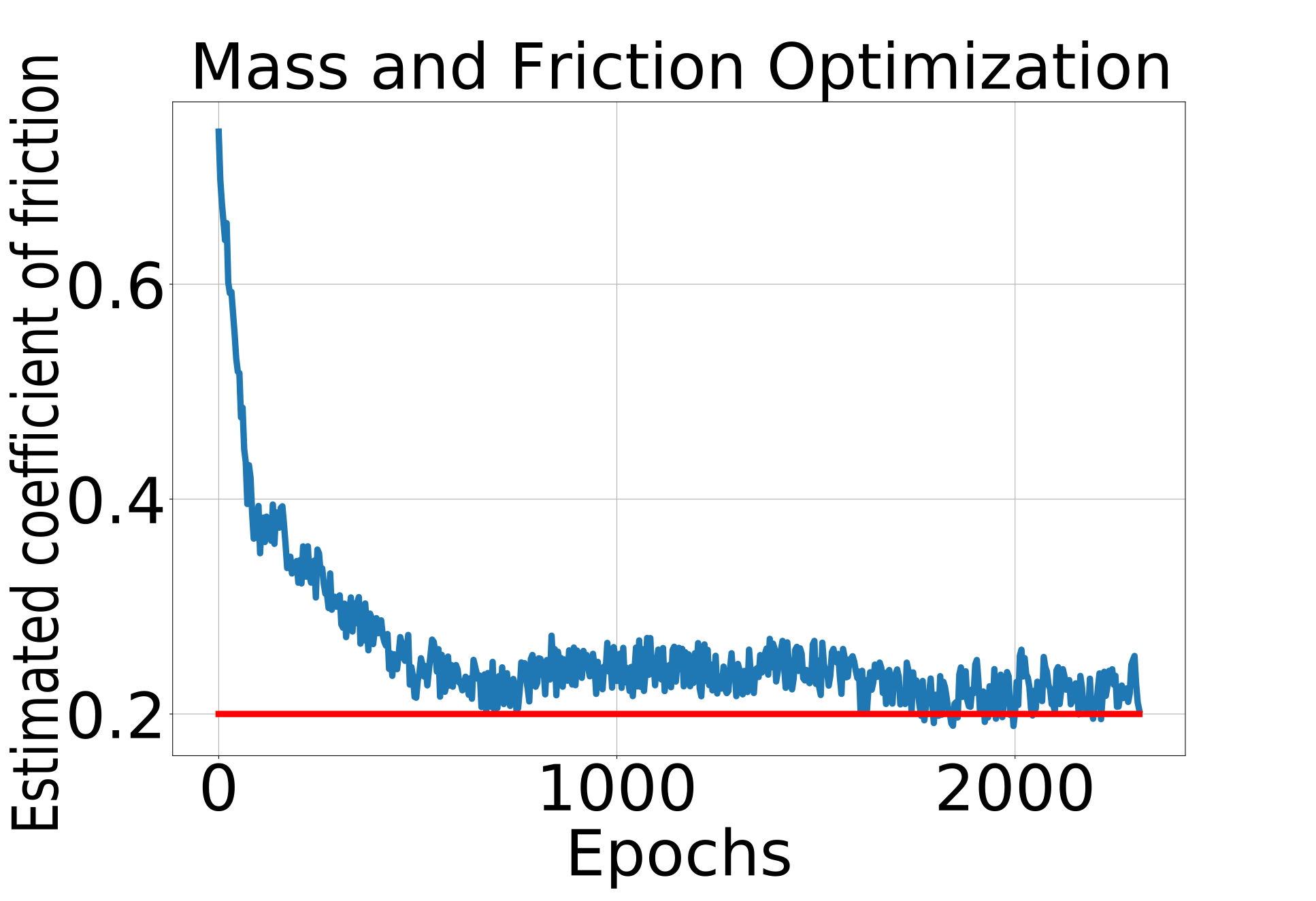} \tiny{rotation inference error: $8^{\circ}$}\\
            \bottomrule
        \end{tabular}
        \caption{Supervised learning results for the 3 scenarios. The physical parameters are well identified (blue lines) close to the ground truth values (red lines). }
        \label{tab:supresults}
	\end{table}

	\subsubsection{System Identification Results}
	As a baseline result, system identification (see Sec.~\ref{sec:sysid}) can be achieved within $200$ epochs with an average position error for all the scenarios between $0.7-1.2\%$.
	The physical parameters reach nominal values with high accuracy.
	Detailed results are given in the supplementary material.

	\begin{table}[tb]
        \centering
        \setlength{\tabcolsep}{0.5em}
        \begin{tabular}{ m{1.5cm} m{3.8cm} m{3.8cm} }
            \toprule
            Inference & Block Pushed On a Flat Plane & Block Colliding With Another Block\\
            \midrule
            Mass &
            \includegraphics[scale=0.07]{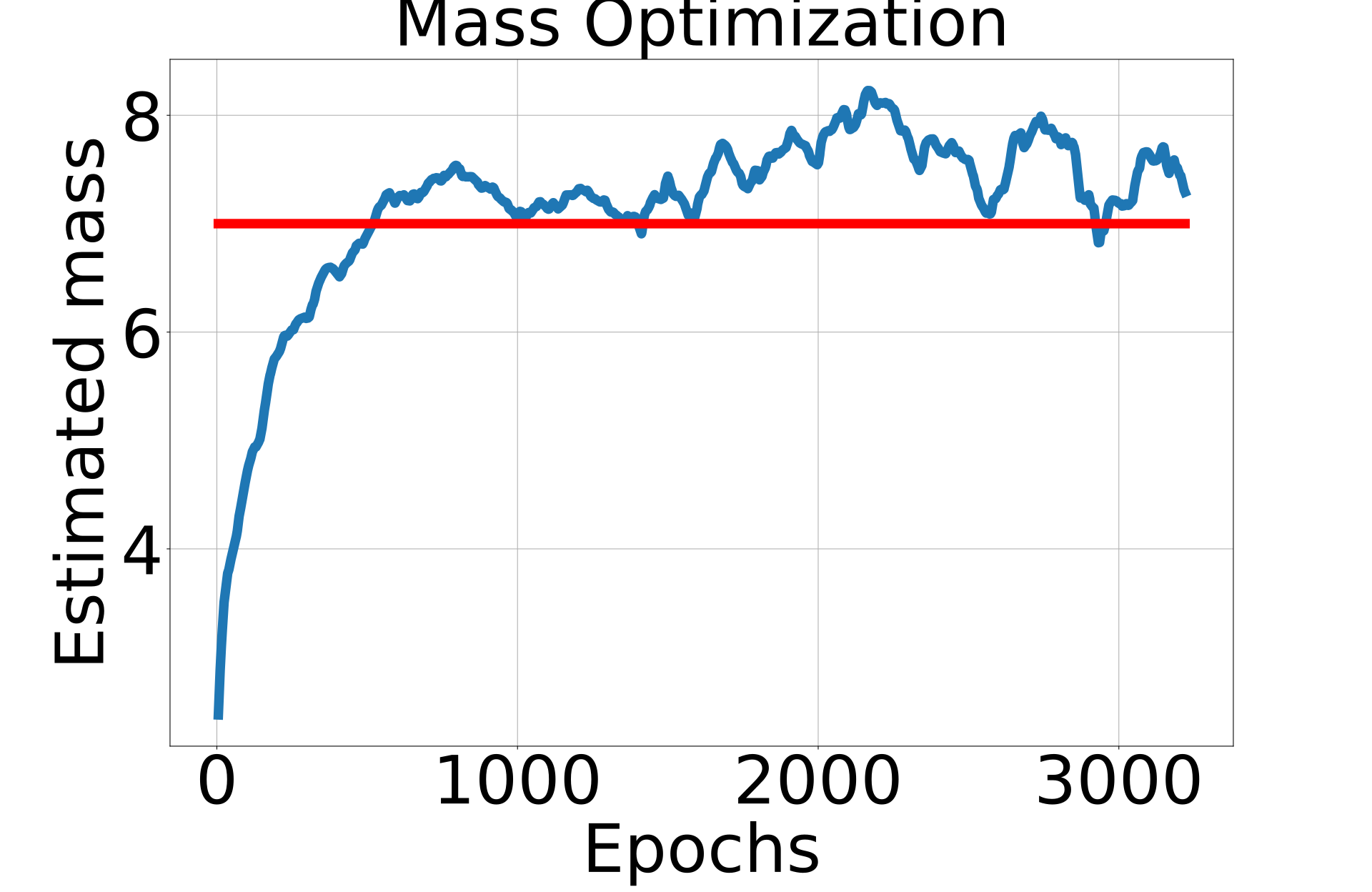} \hspace{2cm}\tiny{position inference error: 7\%}
            & 
            \includegraphics[scale=0.07]{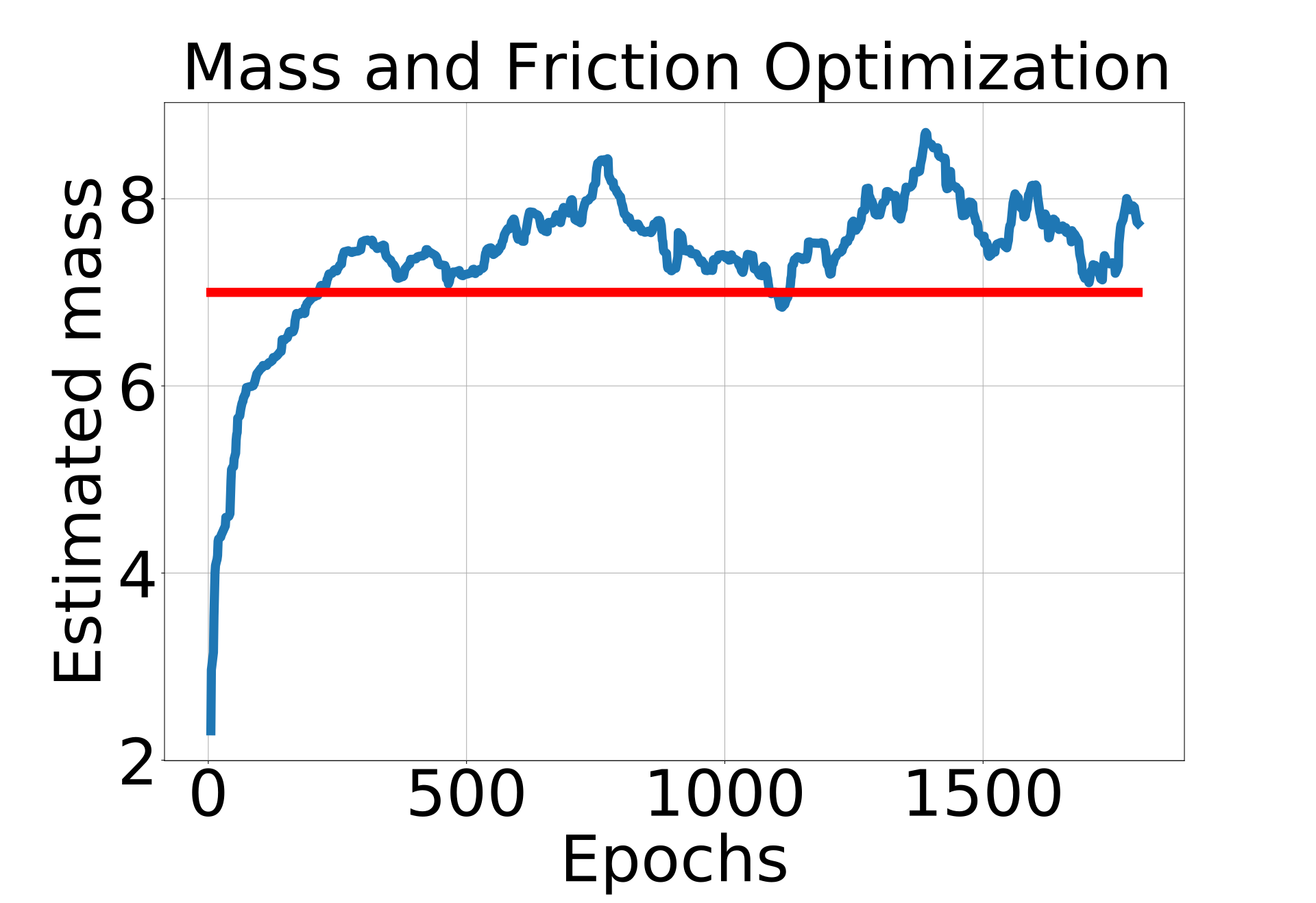} \hspace{2cm} \tiny{position inference error: 12\%}\\
            \midrule
            Coefficient of friction &
            \includegraphics[scale=0.07]{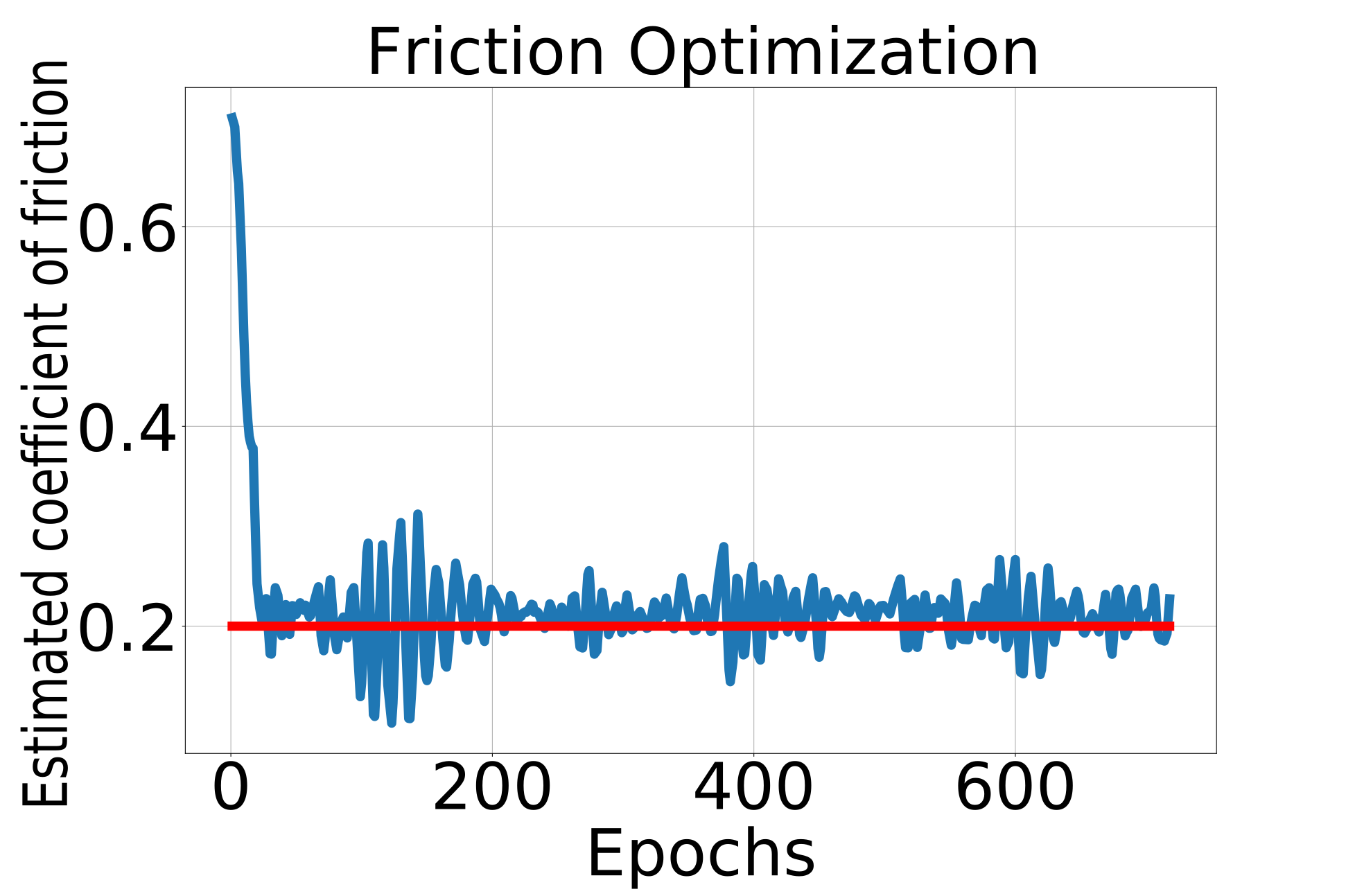} \hspace{2cm} \tiny{position inference error: 8\%}
            & 
            \includegraphics[scale=0.07]{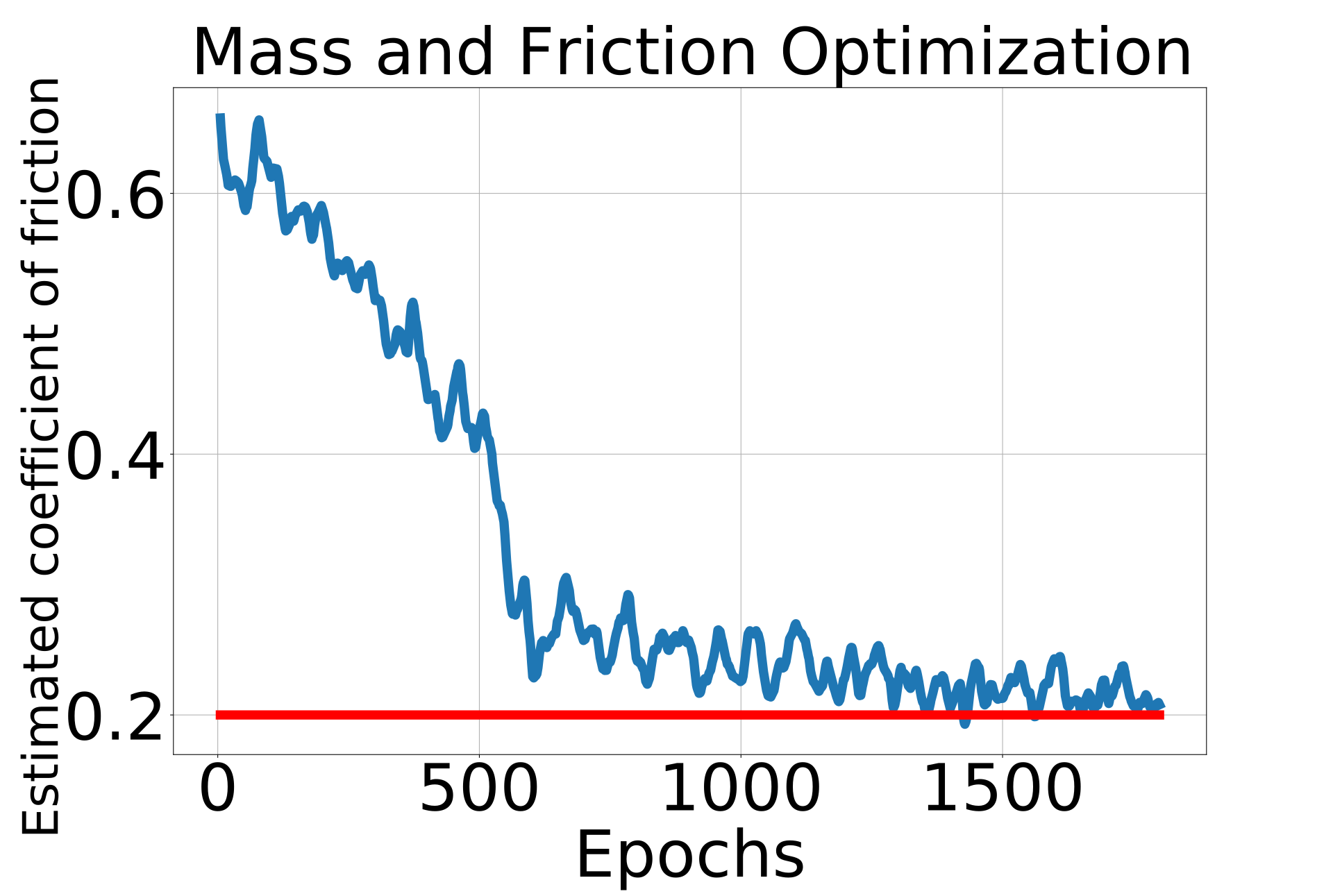} \hspace{2cm} \tiny{rotation inference error: $8^{\circ}$}\\
            \bottomrule
        \end{tabular}
        \caption{Self-supervised learning results for the pushing and collision scenarios. While the encoder error is slightly higher than in the supervised learning case, the physical parameters are identified (blue lines) close to the ground truth values (red lines).}
        \label{tab:selfsupresults}
        \vspace{-2em}
    \end{table}

	\subsubsection{Supervised Learning Results}
	We train our network using the supervised loss in Sec.~\ref{sec:suploss}.
	We warm up the encoder by pre-training with ground truth poses so that when optimizing for physics parameters the training of the encoder is stable.
	From Table~\ref{tab:supresults}, we observe that all the learned physical parameters (in blue) slightly oscillate around the ground truth values (in red). 
	The average inferred position error for all the scenarios is between $2-8\%$ and the average inferred rotation error for the collision scenario is $8^{\circ}$. 
	The parameter learning seems to be robust to this degree of accuracy in the estimated initial states.

	\subsubsection{Self-Supervised Learning Results}
	\label{sec:selfsupresults}
	Now, we train the network in a self-supervised way (see Sec.~\ref{sec:selfsuploss}).
	In this experiment, we generate sequences where the objects start at random locations with zero initial velocity, since the initial velocity estimate is ambiguous for our self-supervised learning approach.
	We obtain average velocities from the estimated poses (Eq.~\eqref{eq: final velocity from poses}).
	Since the pose estimation error is high in self-supervised experiments, the accuracy in velocity especially at the beginning of training is not sufficient for self-supervised learning. 
	We pre-train the encoder in an encoder-decoder way so that when optimizing for physics parameters the training is stable.
	To provide the network with gradients for localizing the objects, we use Gaussian smoothing on the input and reconstructed images starting from kernel size 128 and standard deviation 128, and reducing it to kernel size 5 and standard deviation 2 by the end of training. 
    From Table~\ref{tab:selfsupresults}, we observe that our approach can still recover the physical parameters at good accuracy. 
    Expectably, they are less accurate than in the supervised learning experiment.
    The average inferred position error for all the scenarios is between $7-12\%$ and the average inferred rotation error for the collision scenario is $8^{\circ}$. 
    Through the use of spatial transformers our approach is limited to rendering rectangles in top-down views and cannot handle 3D translation and rotation in our third scenario.

    \subsection{Qualitative Video Prediction Results}

\begin{figure}[tb]
        \centering
        \begin{subfigure}{.5\textwidth}
            \centering
            \includegraphics[width=0.95\linewidth]{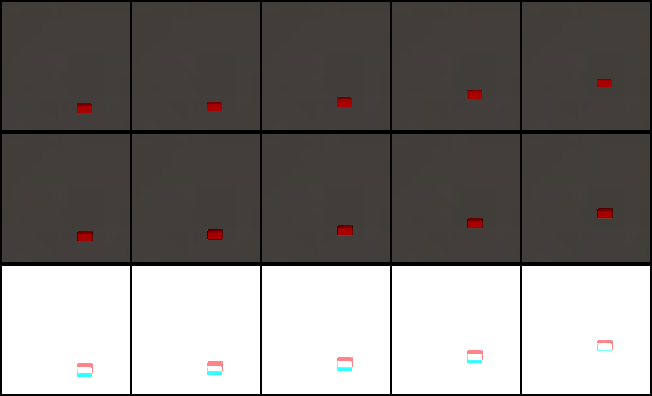}
            \label{fig:sub1}
        \end{subfigure}%
        \begin{subfigure}{.5\textwidth}
            \centering
            \includegraphics[width=0.95\linewidth]{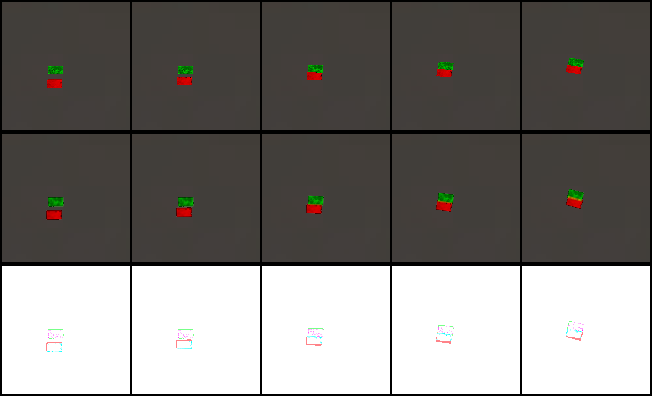}
            \label{fig:sub2}
        \end{subfigure}
        \caption{Qualitative video prediction results for block pushing (left) and collision scenarios (right) with our method. Top: simulated images (from left to right frames 0, 30, 60, 120, 180). Middle: predicted images by our approach. Bottom: difference images.}
        \label{fig:vidpredpushblock}
    \end{figure}

    The learned model in Sec.~\ref{sec:selfsupresults} can be used for video prediction. 
    The images in the top row in Figs.~\ref{fig:vidpredpushblock}(a) and ~\ref{fig:vidpredpushblock}(b) are the ground truth, the images in the middle row are the reconstructions from the predicted trajectories by our network and
    the images in the bottom row are the difference images.
    We roll out a four second trajectory. 
    We can observe that the positions of the objects are well predicted by our approach, while the approach yields small inaccuracies in predicting rotations which occur after the collision of the objects.
    Further video prediction results are included in the supplementary material.
    
    \subsection{Discussion and Limitations}
    
    Our approach achieves good results for supervised and self-supervised learning in the evaluated scenarios.
    We have studied observability and feasibility of learning physical parameters and video embedding by our approach. At its current stage, our architecture makes several assumptions on the scenes which could be addressed in future research.
    Our approach for using 2D spatial transformers for image generation restricts the self-supervised learning approach to known object shape and appearance and top down views. 
    For real scenes our methods needs information about the applied forces which can be obtained from a known dynamics model (e.g. of a robot) or force sensors.
    For self-supervised learning, methods for bridging the sim-to-real domain gap have to be investigated.

	\section{Conclusion}
	\label{conclusion}
	
	In this paper we study supervised and self-supervised learning approaches to learn image encodings and identify physical parameters.
	Our deep neural network architecture integrates differentiable physics with a spatial transformer network layer to learn a physical latent representation of video and applied forces.
    For supervised learning, an encoder regresses the initial object state from images.
    Self-supervised learning is achieved through the implementation of a spatial transformer which decodes the predicted positions by the encoder and the physics engine back into images.
    This way, the model can also be used for video prediction with known actions by letting the physics engine predict positions and velocities conditioned on the actions.
	We evaluate our approach in scenarios which include pushing, sliding and collision of objects. 
	We analyze the observability of physical parameters and assess the quality of the reconstruction of these parameters using our learning approaches.
	In future work we plan to investigate further scenarios including inelastic collisions with restitution and extend our self-supervised approach to real scenes and full 3D motion of objects.
	
	\section*{Acknowledgements}
	We acknowledge support from Cyber Valley, the Max Planck Society, and the German Federal Ministry of Education and Research (BMBF) through the Tuebingen AI Center (FKZ: 01IS18039B). 
	The authors thank the International Max Planck Research School for Intelligent Systems (IMPRS-IS) for supporting Jan Achterhold.

	\bibliographystyle{splncs03}
	\bibliography{egbib}

\begin{thebibliography}{10}
\providecommand{\url}[1]{\texttt{#1}}
\providecommand{\urlprefix}{URL }

\bibitem{optnet}
Amos, B., Kolter, J.Z.: Optnet: Differentiable optimization as a layer in
  neural networks. In: International Conference on Machine Learning. p.
  136–145 (2017)

\bibitem{FrictionModeling}
Anitescu, M., Potra, F.A.: Formulating dynamic multi-rigid-body contact
  problems with friction as solvable linear complementarity problems. Nonlinear
  Dynamics  14,  231--247 (1997)

\bibitem{belbuteperes2018_diffphys}
de~Avila Belbute-Peres, F., Smith, K., Allen, K., Tenenbaum, J., Kolter, J.Z.:
  End-to-end differentiable physics for learning and control. In: Advances in
  Neural Information Processing Systems. pp. 7178--7189 (2018)

\bibitem{babaeizadeh2018_stochvarvideopred}
Babaeizadeh, M., Finn, C., Erhan, D., Campbell, R., Levine, S.: Stochastic
  variational video prediction. In: Proceedings of the International Conference
  on Learning Representations (2018)

\bibitem{chen2018_tcbetavae}
Chen, R.T.Q., Li, X., Grosse, R.B., Duvenaud, D.K.: Isolating sources of
  disentanglement in variational autoencoders. In: Advances in Neural
  Information Processing Systems. pp. 2610--2620 (2018)

\bibitem{elu}
Clevert, D.A., Unterthiner, T., Hochreiter, S.: Fast and accurate deep network
  learning by exponential linear units (elus). In: Proceedings of the
  International Conference on Learning Representations (2016)

\bibitem{Cline_2002}
Cline, M.B.: Rigid body simulation with contact and constraints. Ph.D. thesis
  (2002),
  \url{https://open.library.ubc.ca/collections/ubctheses/831/items/1.0051676}

\bibitem{degrave2019_diffphys}
Degrave, J., Hermans, M., Dambre, J., Wyffels, F.: A differentiable physics
  engine for deep learning in robotics. Frontiers in Neurorobotics  13 (2016)

\bibitem{finn2017_visualforesight}
Finn, C., Levine, S.: Deep visual foresight for planning robot motion. In:
  International Conference on Robotics and Automation. pp. 2786--2793 (2017)

\bibitem{finn2016_unsupvideopredact}
Finn, C., Goodfellow, I.J., Levine, S.: Unsupervised learning for physical
  interaction through video prediction. In: Advances in Neural Information
  Processing Systems. pp. 64--72 (2016)

\bibitem{wallach2019_hamnn}
Greydanus, S., Dzamba, M., Yosinski, J.: Hamiltonian neural networks. In:
  Advances in Neural Information Processing Systems. pp. 15379--15389 (2019)

\bibitem{hafner2019_planet}
Hafner, D., Lillicrap, T., Fischer, I., Villegas, R., Ha, D., Lee, H.,
  Davidson, J.: Learning latent dynamics for planning from pixels. In:
  International Conference on Machine Learning. pp. 2555--2565 (2019)

\bibitem{hochreiter1997_lstm}
Hochreiter, S., Schmidhuber, J.: Long short-term memory. Neural computation  9,
   1735--80 (1997)

\bibitem{stn}
Jaderberg, M., Simonyan, K., Zisserman, A., Kavukcuoglu, K.: Spatial
  transformer networks. In: Advances in Neural Information Processing Systems.
  pp. 2017--2025 (2015)

\bibitem{jaques2020_physasinvgraph}
Jaques, M., Burke, M., Hospedales, T.M.: Physics-as-inverse-graphics: Joint
  unsupervised learning of objects and physics from video. Proceedings of the
  International Conference on Learning Representations  (2020)

\bibitem{vae}
Kingma, D.P., Welling, M.: Auto-encoding variational bayes. In: Proceedings of
  the International Conference on Learning Representations (2014)

\bibitem{analyticalLearned}
Kloss, A., Schaal, S., Bohg, J.: Combining learned and analytical models for
  predicting action effects. CoRR  abs/1710.04102 (2017)

\bibitem{boyd}
Mattingley, J., Boyd, S.: Cvxgen: A code generator for embedded convex
  optimization. Optimization and Engineering  13 (2012)

\bibitem{mottaghi2016_newtonian}
Mottaghi, R., Bagherinezhad, H., Rastegari, M., Farhadi, A.: Newtonian scene
  understanding: Unfolding the dynamics of objects in static images. In:
  Proceedings of the IEEE Conference on Computer Vision and Pattern Recognition
  (2016)

\bibitem{motthagi2016_whatif}
Mottaghi, R., Rastegari, M., Gupta, A., Farhadi, A.: ``{What} happens if...''
  learning to predict the effect of forces in images. In: European Conference
  on Computer Vision (2016)

\bibitem{runia2020_clothinthewind}
Runia, T.F.H., Gavrilyuk, K., Snoek, C.G.M., Smeulders, A.W.M.: Cloth in the
  wind: A case study of estimating physical measurement through simulation. In:
  Proceedings of the IEEE Conference on Computer Vision and Pattern Recognition
  (2020)

\bibitem{shi2015_convlstm}
Shi, X., Chen, Z., Wang, H., Yeung, D.Y., Wong, W.k., Woo, W.c.: Convolutional
  lstm network: A machine learning approach for precipitation nowcasting. In:
  Advances in Neural Information Processing Systems. pp. 802--810 (2015)

\bibitem{srivastava2015_unsupvideoreplstm}
Srivastava, N., Mansimov, E., Salakhutdinov, R.: Unsupervised learning of video
  representations using lstms. In: International Conference on Machine Learning
  (2015)

\bibitem{FrictionModeling2}
Stewart, D.: Rigid-body dynamics with friction and impact. SIAM Rev.  42,
  3--39 (2000)

\bibitem{watters2017_vin}
Watters, N., Zoran, D., Weber, T., Battaglia, P., Pascanu, R., Tacchetti, A.:
  Visual interaction networks: Learning a physics simulator from video. In:
  Advances in Neural Information Processing Systems (2017)

\bibitem{ye2018_interppred}
Ye, T., Wang, X., Davidson, J., Gupta, A.: Interpretable intuitive physics
  model. In: European Conference on Computer Vision (2018)

\bibitem{zhu2019_physvidpred}
Zhu, D., Munderloh, M., Rosenhahn, B., St{\"u}ckler, J.: Learning to
  disentangle latent physical factors for video prediction. In: German
  Conference on Pattern Recognition (2019)

\end{thebibliography}

	\newpage
	\flushleft{\textbf{\Large Supplementary Material}}
	\vspace{0.5cm}

	\setcounter{equation}{0}
	\setcounter{figure}{0}
	\setcounter{table}{0}
	\setcounter{section}{0}
	
	\renewcommand{\thesection}{S\arabic{section}}
	\renewcommand{\theequation}{\thesection.\arabic{equation}}
	\renewcommand{\thefigure}{\thesection.\arabic{figure}}
	\renewcommand{\thetable}{\thesection.\arabic{table}}
	
	\noindent
	In the following we provide further details on the architecture and training pipelines of our approach. 
	We also present system identification and additional qualitative results for learning friction in the pushing and collision scenarios using our self-supervised learning approach.
	
	\section{Encoder Architecture}
    We provide further details of our network architectures for the supervised and self-supervised learning approaches in Tables~\ref{tab:enc_convs} and~\ref{tab:enc_fcs}.
	
	\begin{table}[h!]
        \centering
        \setlength{\tabcolsep}{0.5em}
        \begin{tabular}{ccccccc}
            \toprule
            \textbf{Layer} & \textbf{Type} & \textbf{Input channels} & \textbf{Output channels} & \textbf{Kernel} & \textbf{Stride}\\
            \midrule
            1 & Conv & 4 & 64 & 5 & 2\\
            2 & Conv & 64 & 128 & 5 & 2\\
            3 & Conv & 128 & 256 & 5 & 2\\
            4 & Conv & 256 & 256 & 5 & 2\\
            5 & Conv & 256 & 128 & 3 & 1\\
            \bottomrule
        \end{tabular}
	\caption{Convolutional layers in our encoder architecture.}
	\label{tab:enc_convs}
    \end{table}
    \begin{table}[h!]
        \centering
        \setlength{\tabcolsep}{0.5em}
        \begin{tabular}{cccc}
            \toprule
            \textbf{Layer} & \textbf{Type} & \textbf{Input Size} & \textbf{Output Size} \\
            \midrule
            5 & FC & 3*3*128 & 50\\ 
            6 & FC & 50 & 50\\ 
            7 & FC & 50 & latent size\\
            \bottomrule
        \end{tabular}
	\caption{Fully connected layers in our encoder architecture.}
	\label{tab:enc_fcs}
    \end{table}
    
    \section{Decoder Architecture}
    
    To train the pipeline in a self-supervised way, we need a decoder which interprets the output of the physics engine layer and renders the objects at the estimated poses. 
    For that purpose, we use a spatial transformer network (STN~\cite{stn}) layer. 
    Since we use the STN layer only for rendering, we do not have any learnable parameters for the decoder.
    
    We use the absolute poses predicted by the physics engine layer to render the images.
    We assume known shape and appearance of the object and extract a content image for each object from the first image of the sequence using ground-truth segmentation masks.
    The predicted poses are converted to image positions and in-plane rotations of the rectangular shape of the object in the top-down view assuming known camera intrinsics and extrinsics.
    A spatial transformer network layer renders the object's content image at the predicted image position and rotation. 
    Finally, the prediction is reconstructed by overlaying the transformed object images onto the known background.
    
	\section{Training Pipelines}
\begin{figure}[hb!]
    \centering
        \includegraphics[scale=0.43]{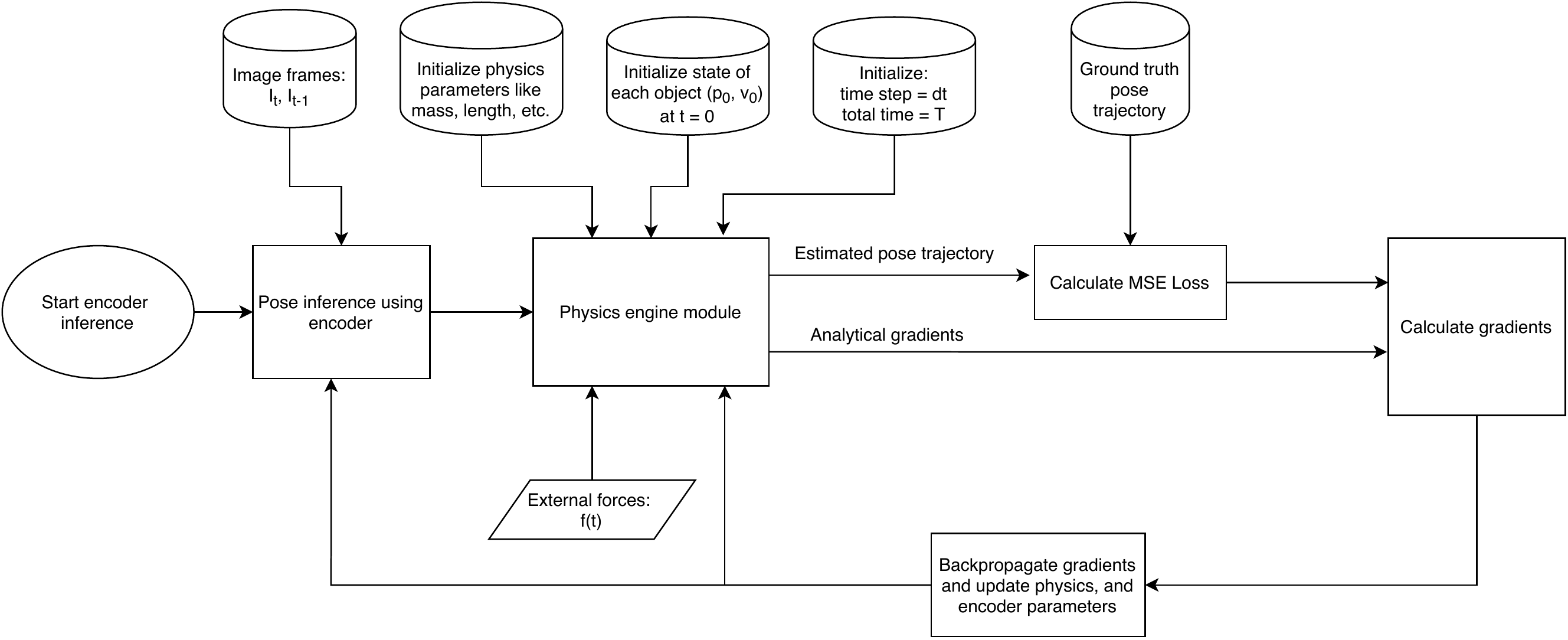}
        \caption{Supervised learning pipeline}
        \label{fig: Supervised learning with encoder pipeline}
    \end{figure}

\begin{figure}[hb]
    \centering
        \includegraphics[scale=0.43]{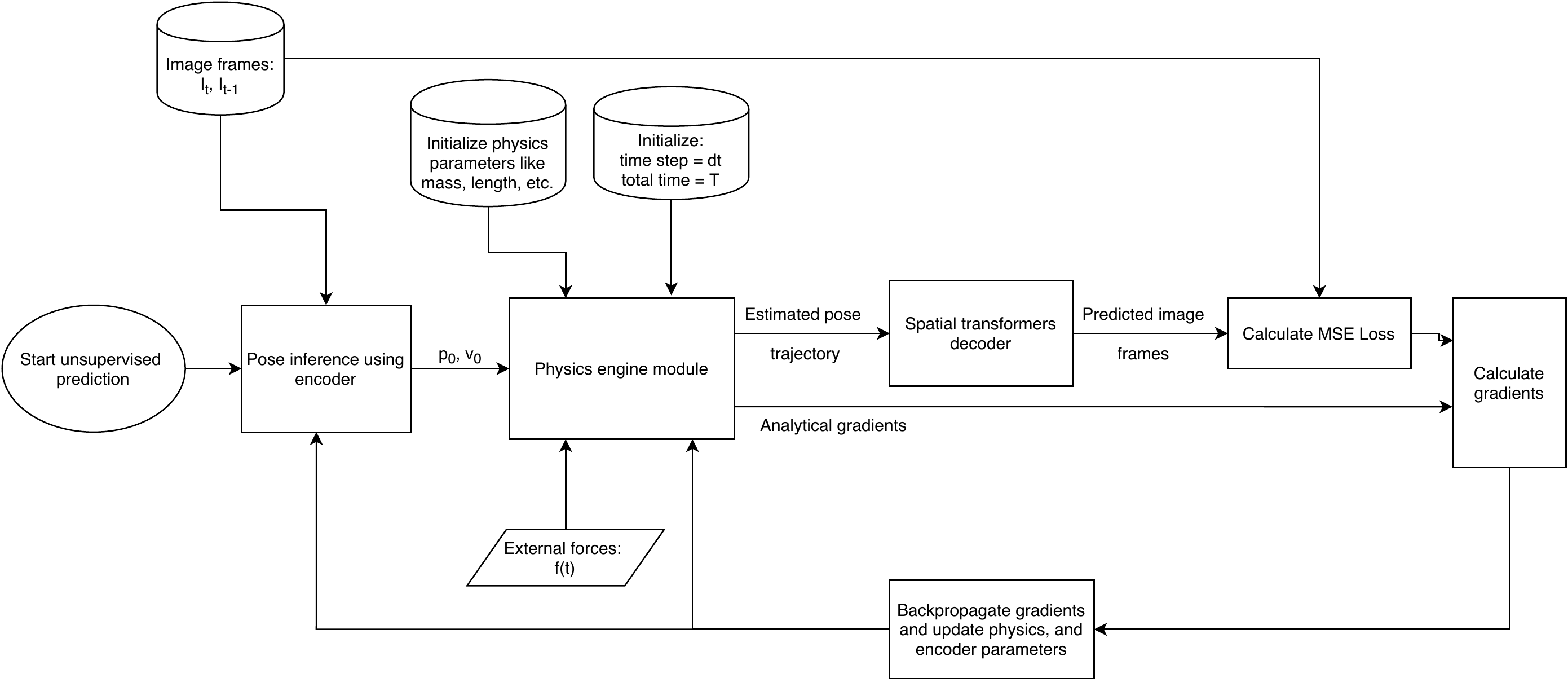}
        \caption{Self-supervised learning pipeline}
        \label{fig: self supervised learning with encoder pipeline}
    \end{figure}

\begin{figure}[ht]
    \centering
        \includegraphics[scale=0.5]{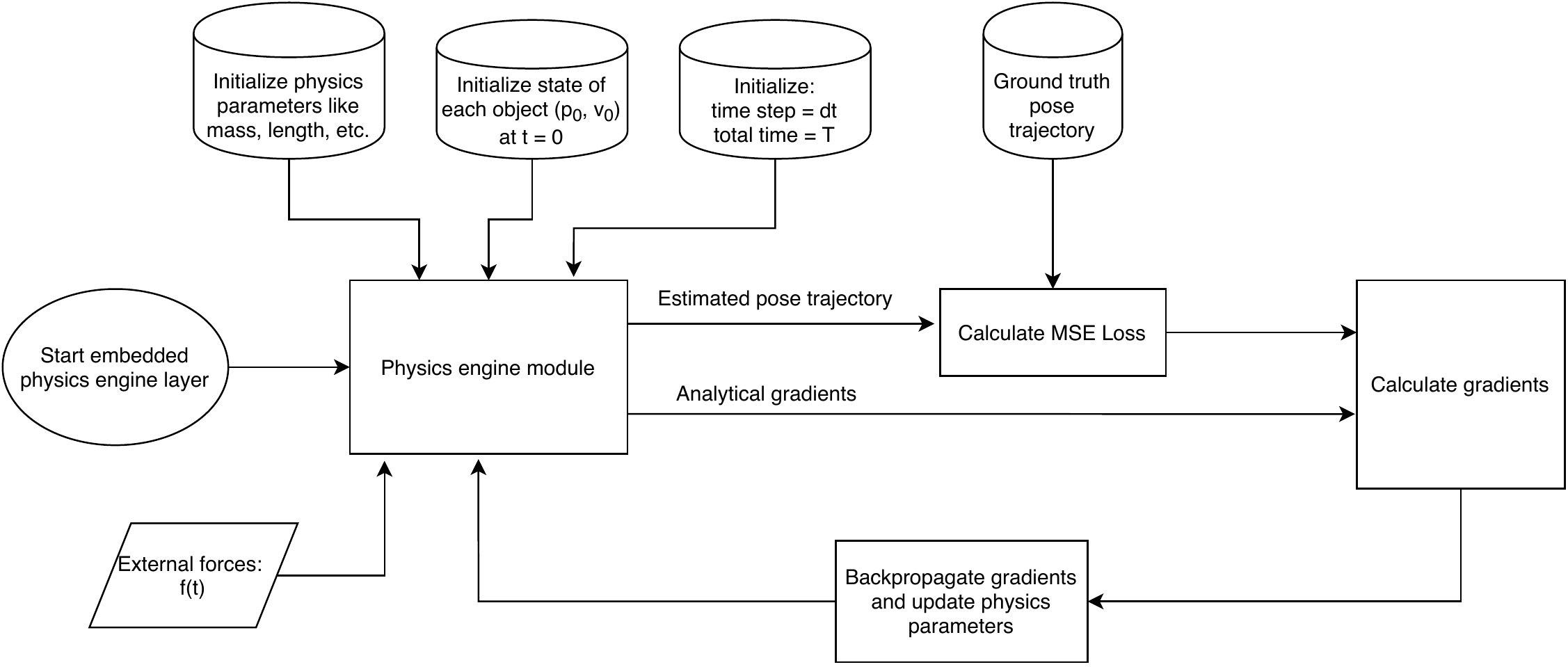}
        \caption{System identification pipeline}
        \label{fig: system identification pipeline}
    \end{figure}

\begin{figure}[hb]
        \centering
        \includegraphics[scale=0.61]{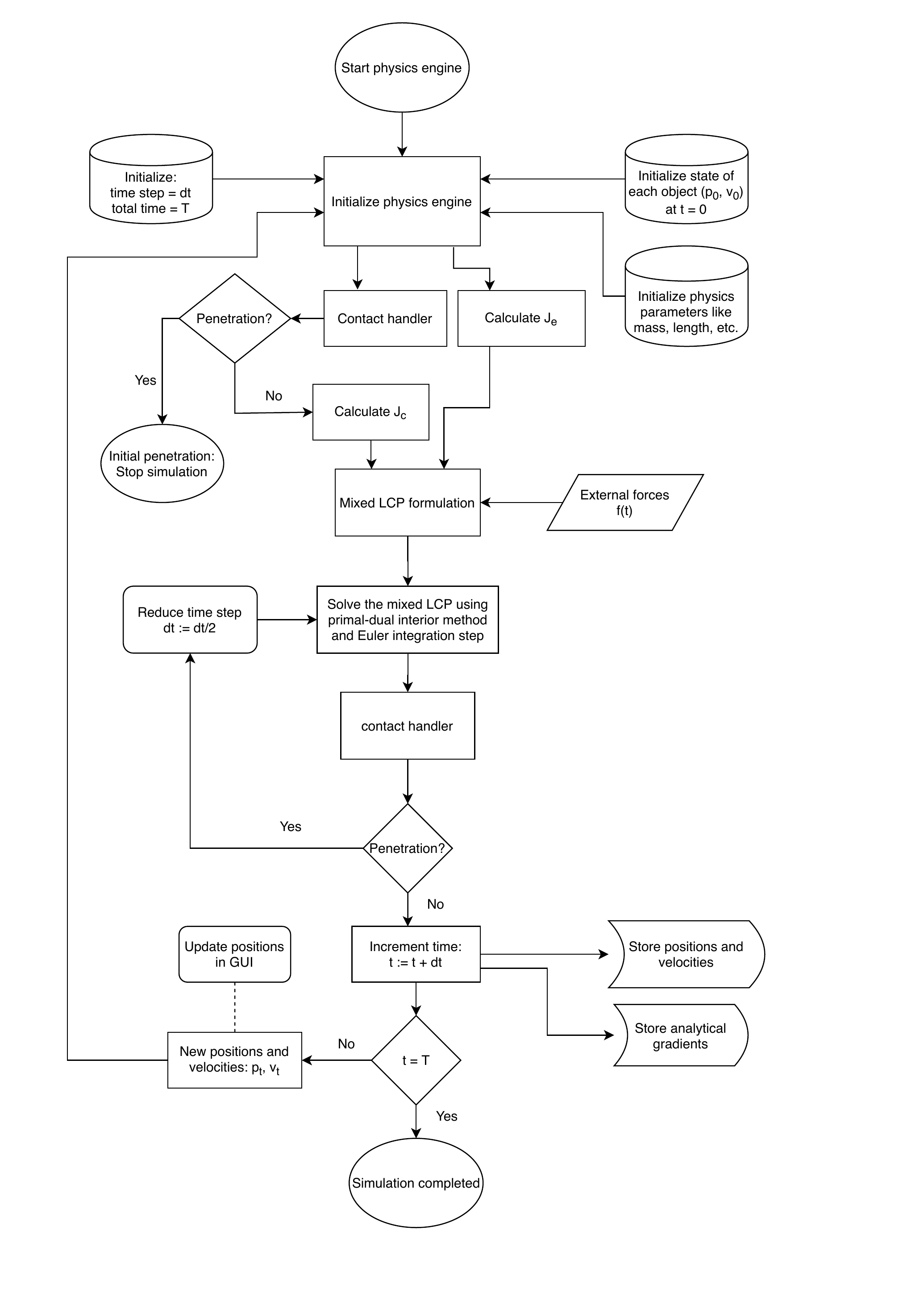}
        \caption{Differentiable physics engine module}
        \label{fig: differentiable physics engine module}
    \end{figure}

	We provide flow diagrams to visualize our training pipelines for supervised (Fig.~\ref{fig: Supervised learning with encoder pipeline}), self-supervised learning (Fig.~\ref{fig: self supervised learning with encoder pipeline}) and system identification (Fig.~\ref{fig: system identification pipeline}).
    Fig.~\ref{fig: differentiable physics engine module} illustrates the processing steps of the differentiable physics engine as described in~\cite{belbuteperes2018_diffphys}.
    
\clearpage
	\section{Additional Results}

	\subsection{System Identification Results}
	Table~\ref{tab:sysidresults} plots the estimates of the physical parameters for system identification.
	Notably, the parameters quickly convergence in about 200 epochs close to the ground truth values.

    \begin{table}[h!]
        \center
        \setlength{\tabcolsep}{0.5em}
        \begin{tabular}{m{2cm}m{3cm}m{3cm}m{3cm}}
            \toprule
            Inference & Block Pushed On a Flat Plane & Block Sliding Down the Inclined Plane & Block Colliding With Another Block\\
            \midrule
            Mass &
            \includegraphics[scale=0.065]{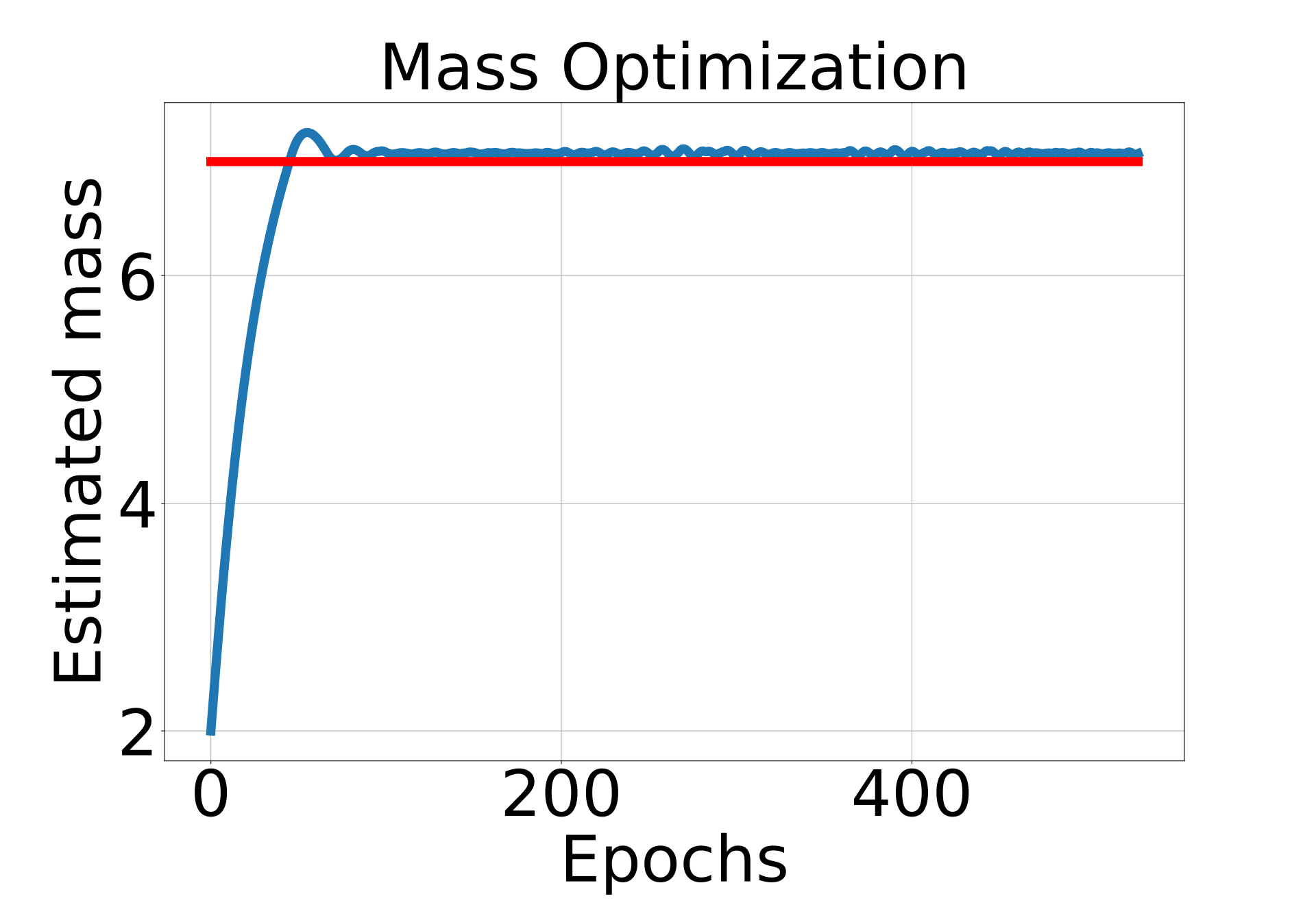} \tiny{position trajectory error: 0.7\%}
            &
            \centering Not feasible
            & 
            \includegraphics[scale=0.065]{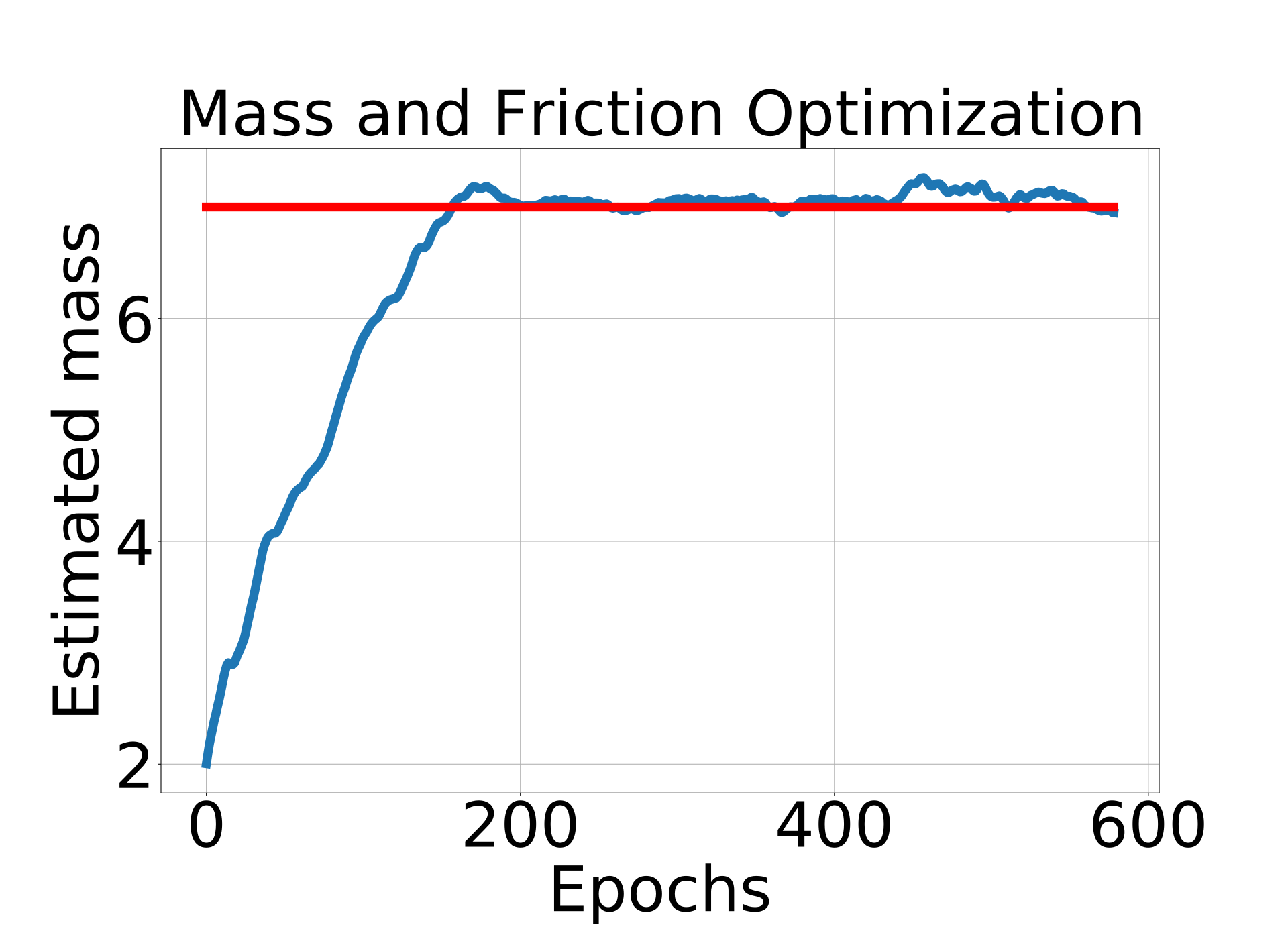} \tiny{position trajectory error: 1.2\%}\\
            \midrule
            Coefficient of friction &
            \includegraphics[scale=0.065]{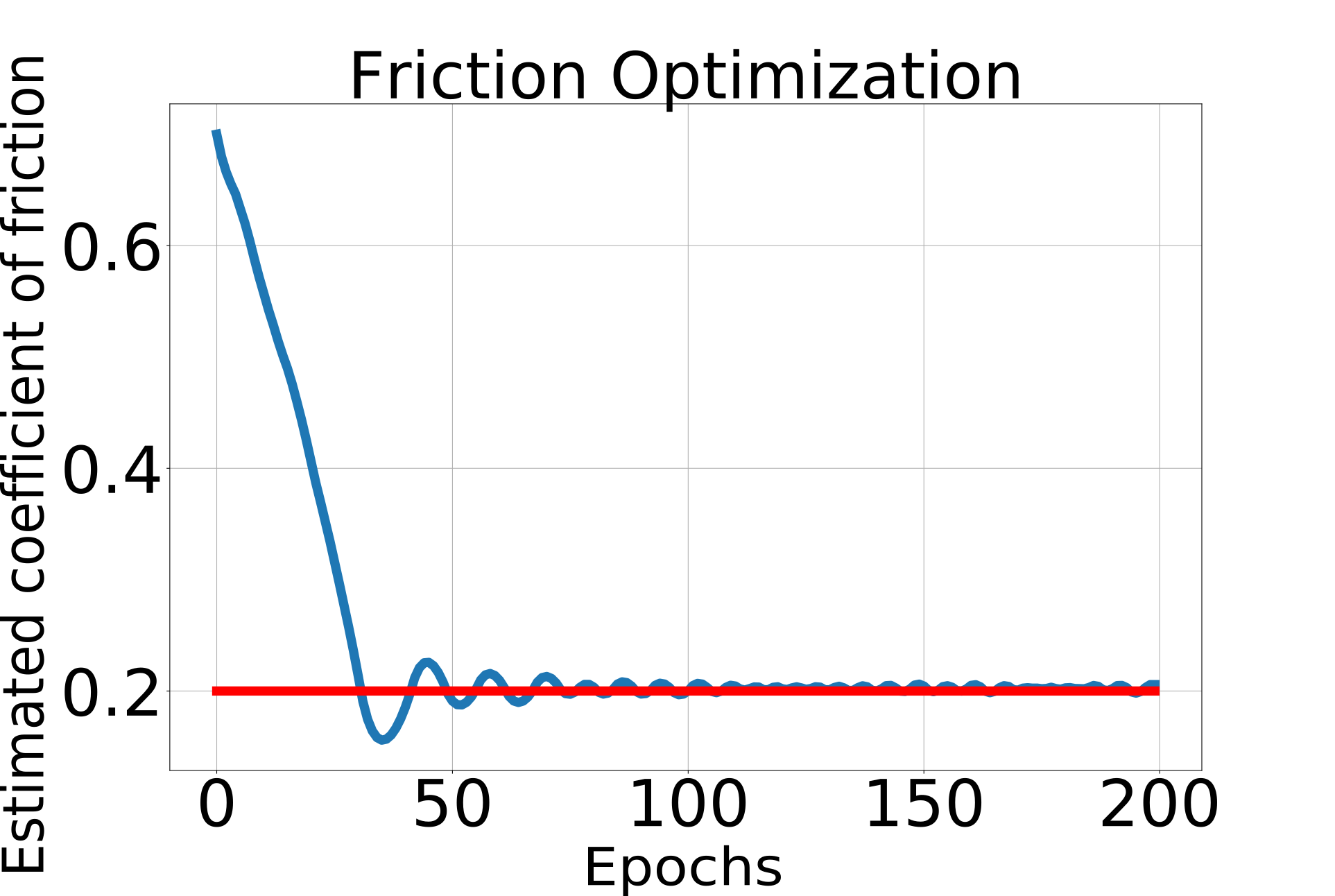} \tiny{position trajectory error: 0.5\%}
            &
           \includegraphics[scale=0.065]{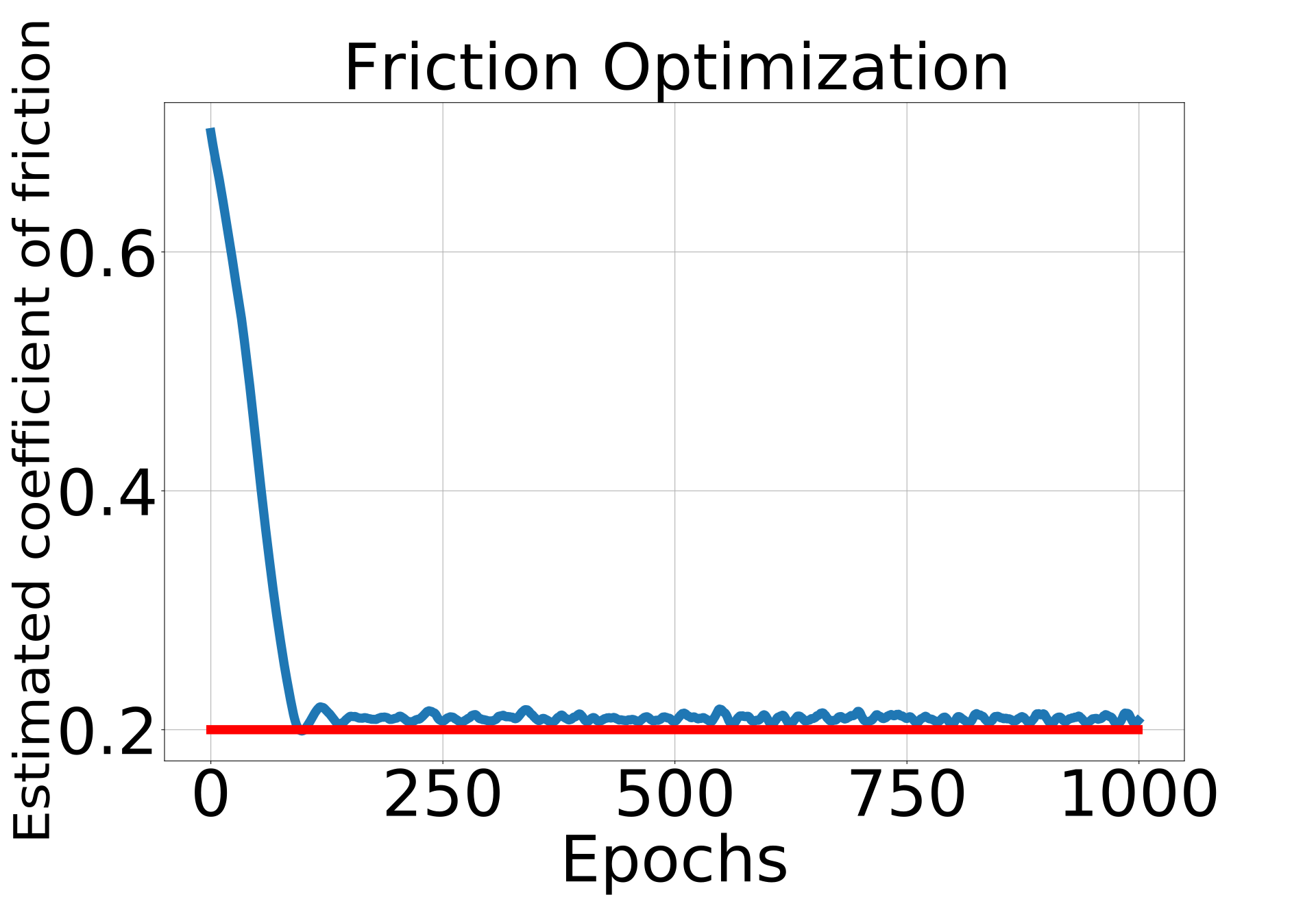} \tiny{position trajectory error: 0.7\%}
            & 
            \includegraphics[scale=0.065]{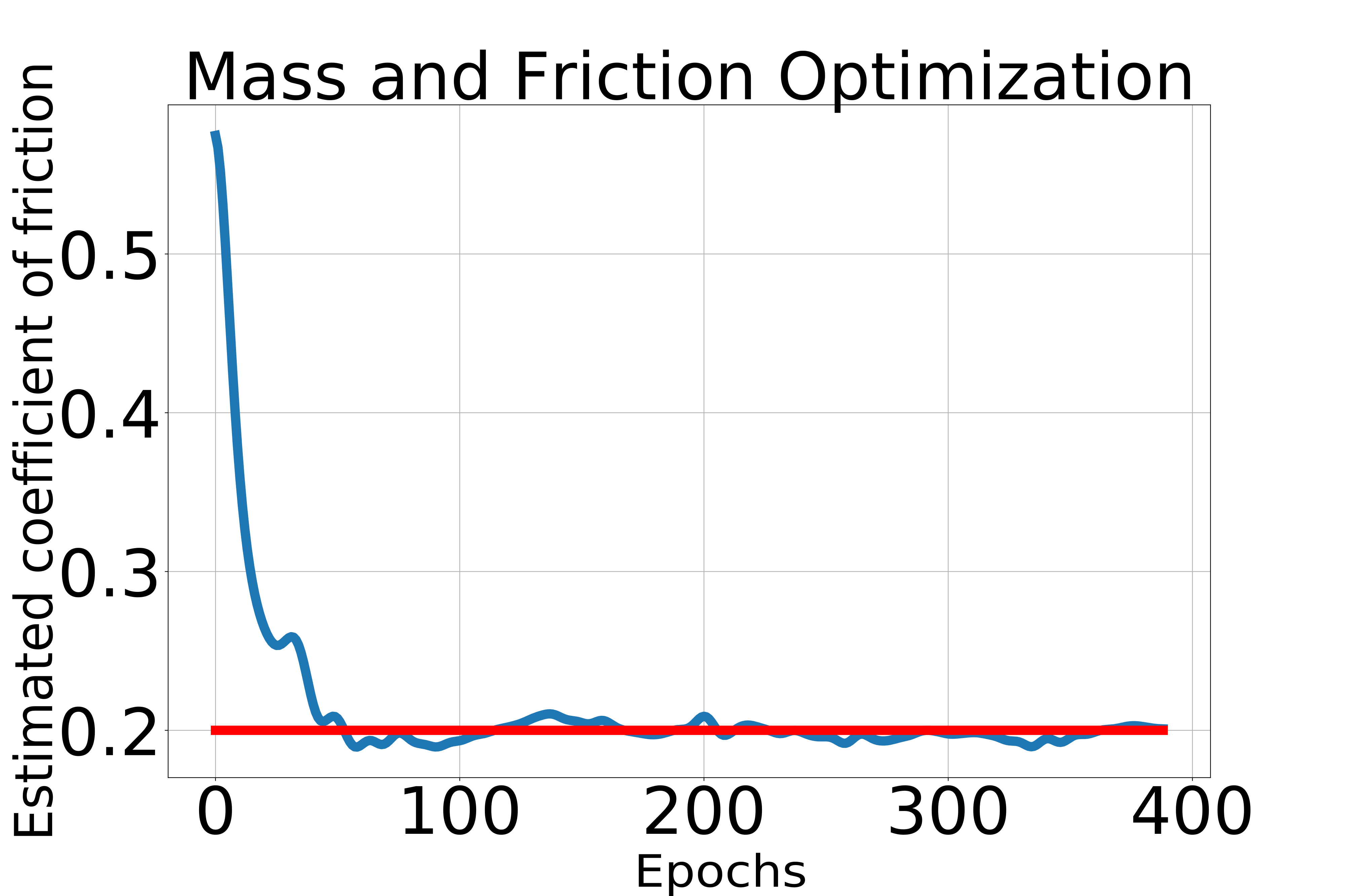}\\
            \bottomrule
        \end{tabular}
        \caption{System identification results (blue lines) for the 3 scenarios. Ground truth parameters are shown as red lines.}
        \label{tab:sysidresults}
	\end{table}

    \subsection{Qualitative Video Prediction Results}
\begin{figure}[h!]
        \centering
        \includegraphics[scale=0.33]{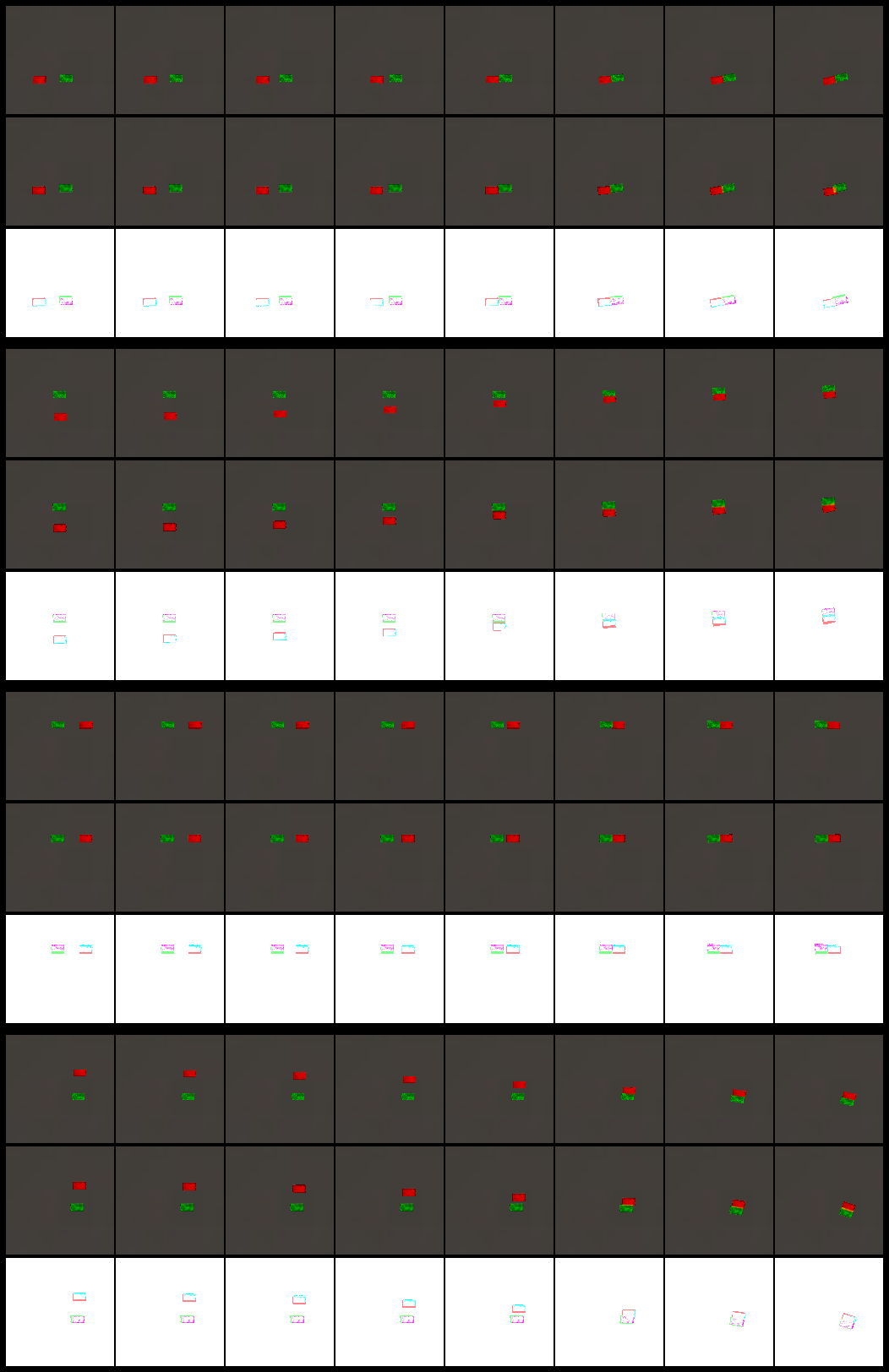}
        \caption{Video prediction results by our self-supervised learning approach based on the first frame after learning friction for the pushing scenario. First row: simulated images. Second row: predicted video frames. Third row: Difference images between ground truth and reconstructed images.}
        \label{fig:vidpred}
    \end{figure}

    In Fig.~\ref{fig:vidpred} we show additional samples of qualitative video prediction results based on our self-supervised learning approach.

\end{document}